\documentclass[11pt, a4paper]{article}
 \usepackage{fullpage}
 
\usepackage{natbib}
\renewcommand{\bibname}{References}
\renewcommand{\bibsection}{\subsubsection*{\bibname}}

\usepackage{amsmath,amsfonts,bm}

\def\eqref#1{equation~\ref{#1}}

\def\1{\bm{1}}

\def\ve{{\bm{e}}}
\def\vf{{\bm{f}}}

\def\vq{{\bm{q}}}

\def\vw{{\bm{w}}}
\def\vx{{\bm{x}}}
\def\vy{{\bm{y}}}
\def\vz{{\bm{z}}}

\def\mA{{\bm{A}}}
\def\mB{{\bm{B}}}
\def\mC{{\bm{C}}}
\def\mD{{\bm{D}}}
\def\mE{{\bm{E}}}
\def\mF{{\bm{F}}}
\def\mG{{\bm{G}}}
\def\mH{{\bm{H}}}
\def\mI{{\bm{I}}}
\def\mJ{{\bm{J}}}

\def\mL{{\bm{L}}}

\def\mQ{{\bm{Q}}}
\def\mR{{\bm{R}}}

\def\mT{{\bm{T}}}
\def\mU{{\bm{U}}}
\def\mV{{\bm{V}}}
\def\mW{{\bm{W}}}
\def\mX{{\bm{X}}}

\DeclareMathAlphabet{\mathsfit}{\encodingdefault}{\sfdefault}{m}{sl}
\SetMathAlphabet{\mathsfit}{bold}{\encodingdefault}{\sfdefault}{bx}{n}

\newcommand{\E}{\mathbb{E}}

\newcommand{\R}{\mathbb{R}}

\usepackage{amsmath,amssymb,amsthm}
\usepackage{hyperref}
\usepackage{url}
\usepackage{booktabs,adjustbox,caption}
\usepackage{longtable}
\usepackage{tabularx}
\usepackage{diagbox}
\usepackage{xcolor}
\usepackage{multirow}
\usepackage{titlesec}

\titleformat*{\section}{\Large\bfseries\sffamily}
\titleformat*{\subsection}{\large\bfseries\sffamily}

\newcommand{\tr}{\operatorname{tr}}

\newcommand{\vdelta}{\bm{\delta}}

\newtheorem{assumption}{Assumption}

\newtheorem{proposition}{Proposition}

\providecommand{\keywords}[1]
{
  \small	
  \textbf{{Keywords---}} #1
}

\title{\rm Revisiting Error Feedback for Model Quantization}
\author{Wai-Yiu Keung$^{\dagger}$\thanks{
Equal contribution.}, 
Chi Cui$^{\S}$\footnotemark[1], 
Wing-Kin Ma$^{\dagger}$, 
Jesús Cid-Sueiro$^{\S}$ 
and  
Hei Victor Cheng$^{\ddagger}$\thanks{The corresponding author. Email: \texttt{<hvc@ece.au.dk>}.}
}
\date{
\normalsize $^{\dagger}$The Chinese University of Hong Kong, Hong Kong SAR of China\\
\normalsize $^{\S}$Universidad Carlos III de Madrid, Spain\\
\normalsize $^{\ddagger}$Aarhus University, Denmark
}

\makeatletter
\renewcommand*{\@fnsymbol}[1]{\ensuremath{\ifcase#1\or 
  \ast \or                     
  \star \or                    
  \ast\ast \or              
  \dagger\dagger            
  \else\@ctrerr\fi}}
\makeatother

\begin{document}
\maketitle

\begin{abstract} 
We study causal error-feedback coefficients as design variables for model quantization. 
Modeling instantaneous quantization errors as independent, zero-mean variables with a common fixed variance,
we formulated a metric-dependent trace problem for designing feedback coefficients. 
The formulation resembles fixed-order GPTQ/LDLQ feedback 
and supports Jacobian-aware designs that account for downstream error propagation.
For a positive-definite metric $\mH$, 
we prove that the joint rotation-feedback trace minimum is $n\det(\mH)^{1/n}$
and present an attaining pair using the geometric mean decomposition. 
Experiments on synthetic networks, trained MLPs, and language models show benefits from downstream-sensitive feedback, but also reveal that trace optimality need not improve finite-level performance. 
Small QAT experiments further illustrate direct loss-driven feedback tuning with fixed or trainable network weights.
\end{abstract}


\section{Introduction}
\label{sec:introduction}
Post-training quantization (PTQ) reduces 
the precision of a trained network using limited calibration data, 
without end-to-end retraining~\citep{frantar2022obc,frantar2023gptq}. 
Earlier work on model
compression combined pruning, quantization, and coding,
while PTQ methods developed techniques that do not
require full retraining
~\citep{han2016deepcompression,banner2019aciq,nagel2019datafree}.
AdaRound and BRECQ further established data-dependent rounding and
block reconstruction for PTQ
~\citep{nagel2020adaround,li2021brecq}.
PTQ performance depends not only on the quantization resolution,
but also on how the errors introduced at individual weight coordinates
affect subsequent quantization decisions and the network output. 
In a sequential quantizer, 
an error made at one coordinate can be fed into
the inputs presented to the quantizer at later coordinates. 
We study the coefficients 
governing this causal error feedback as design
variables of the quantizer.

Feedback and adaptive rounding already play 
an important role in PTQ. 
GPTQ and LDLQ use curvature-informed compensation, 
while QuIP combines adaptive rounding with orthogonal preprocessing
~\citep{frantar2023gptq,chee2023quip}. 
SmoothQuant and AWQ use
activation information and
rescaling in quantizer design
~\citep{xiao2023smoothquant,lin2024awq}. QuIP\#, QuaRot, and
SpinQuant further illustrate the use of structured, random, or
learned transformations
~\citep{tseng2024quipsharp,ashkboos2024quarot,liu2025spinquant}.
Other methods change the criterion used to assess quantization error:
QEP accounts for errors propagated from earlier layers, 
GuidedQuant incorporates end-loss gradients, 
and YAQA uses an approximation to end-to-end curvature
~\citep{arai2025qep,kim2025guidedquant,tseng2026yaqa}.
Qronos develops correction and diffusion of quantization errors,
while GPTQ-2D studies adaptive rounding under a two-sided quadratic
metric~\citep{zhang2025qronos,chen2026gptq2d}.
QuIP already formulates adaptive rounding with linear feedback and characterizes its choice under a layer-wise quadratic criterion. YAQA, in turn, develops adaptive rounding using an approximation to end-to-end
curvature. 
Our question is different: within a fixed causal quantizer, 
how can its feedback coefficients be designed under different input-coordinate criteria, 
jointly with a change of representation, or directly from the quantized network's loss?

Our starting point is a causal quantizer whose feedback coefficients form a unit lower-triangular matrix $\mG$. 
Under a fixed isotropic second-moment model for those errors, 
a specified input-coordinate metric yields a tractable trace design problem. 
The input covariance lead to the
feedback structure associated with fixed-order GPTQ/LDLQ.
Replacing that metric with a Jacobian-derived
input-coordinate surrogate gives a downstream-sensitive
feedback design. This surrogate provides a tractable way 
to incorporate downstream sensitivity 
into feedback design without 
directly optimizing network-output error.

We then allow $\mG$ to be designed jointly with an
orthogonal rotation $\mR$. For a specified positive-definite
metric, we characterize the global optimum of the resulting
fixed-statistics trace problem and construct an attaining
rotation-feedback pair using the QRS/geometric mean
decomposition~\citep{zhang2005QRS, jiang2005gmd}. 
Unlike using a rotation
solely as preprocessing, 
this result identifies the best
rotation and causal feedback \emph{together} for the 
trace objective. It provides a benchmark for asking whether
a trace-optimal design 
also performs well after finite-level
quantization.

Our main contributions are as follows:
\begin{itemize}\vspace{-0pt}
\itemsep0em

\item We formulate causal error feedback as an explicit
quantizer-design variable under a general input-coordinate metric.
The resulting trace framework resembles covariance-based
GPTQ/LDLQ feedback as a special case and enables
downstream-sensitive designs through Jacobian-derived metrics.
\item We characterize the global optimum of the joint
rotation-feedback problem for any positive-definite metric,
and construct a pair using the geometric mean
decomposition attaining the lower bound. This result provides a fixed-statistics benchmark for how far rotation and causal feedback can reduce the trace
surrogate.

\item Our experiments expose the gap between optimizing error propagation
and controlling the errors produced by a finite-level quantizer. Experiments show that objective-aware feedback
can improve downstream fidelity, while the joint trace optimum need not
yield better realized performance. Small QAT experiments further show that
feedback coefficients can be optimized directly against the network loss.

\end{itemize}\vspace{-0pt}

\section{Preliminaries}
\label{sec:background}
Consider an $L$-layer neural network parameterized by weights $\{\mW_\ell\}_{\ell=1}^{L}$. For an input data set $\mX$, the network output is
\[
    {\cal F}_{\{\mW_\ell\}}(\mX)
    =\sigma_L\bigl(\mW_L\sigma_{L-1}(\mW_{L-1}\cdots
    \sigma_1(\mW_1\mX)\cdots)\bigr),
\]
where $\sigma_\ell(\cdot)$ is the per-layer activation function, and $\mW_\ell$ is the weight matrix of the network at layer $\ell$.
Here, $\mW_\ell\in\R^{d_\ell\times d_{\ell-1}}$ and $\mX_\ell\in\R^{d_{\ell-1}\times N}$, where the columns of $\mX_\ell$ contain the activation inputs to layer $\ell$.
The storage requirement of the neural network ${\cal F}_{\{\mW_\ell\}}(\cdot)$ increases as the width of each layer grows.
To fully represent ${\cal F}$, one needs to store the matrices $\{\mW_\ell\}_{\ell=1}^{L}$ in high-precision.
In practical computing, the weight matrices are usually represented by 32-bit floating-point numbers (FP32).
This incurs an implementation barrier for neural networks to be used in devices with limited memory and computational power, such as edge devices.

The goal of post-training quantization is to replace the full-precision matrices $\mW_\ell$ with low-bit quantized matrices $\widehat{\mW}_\ell$, without retraining ${\cal F}$ with data labels.
A natural scheme for such a layer-wise quantization task is to replace $\mW_\ell$ by
\[
    \widehat{\mW}_\ell=\mathcal{Q}_M(\mW_\ell)=\mW_\ell+\mQ_\ell,
\]
where $\mathcal{Q}_M$ is the quantization function, $M$ is the number of quantization levels, and $\mQ_\ell$ is the additive weight-error matrix.
Depending on the magnitudes of the entries in $\mW_\ell$, the quantization alphabet can be adjusted.
This is also known as the round-to-nearest (RTN) method in the literature.

\paragraph{GPTQ/OBS-Style Quantization.}
The shortcoming of such a direct quantization scheme is that the quantization error $\mQ_\ell$ can be large and its effect may dominate.
Consider the activation input at a single layer when the direct quantization scheme is used. Denote $\mX_\ell$ as the activated input at layer $\ell$.
The product
\[
    \widehat{\mW}_\ell\mX_\ell
    =\mW_\ell\mX_\ell+\mQ_\ell\mX_\ell
\]
describes the effect of quantization.
This reveals an objective of minimizing $\|\mQ_\ell\mX_\ell\|_F^2$, where $\mX_\ell$ is the data that cannot be modified, whereas $\mQ_\ell$ depends on input to the quantizer $\mathcal{Q}_M(\cdot)$.
The Optimum Brain Surgery/Generalized Post-Training Quantization (OBS/GPTQ) family of quantizers leverages this insight \citep{hassibi1992obs, frantar2023gptq}.

Consider the quantization of a single row in $\mW_\ell$, represented by $\vw^\top$, with quantized vector $\widehat{\vw}$.
The distortion at the layer output is
\[
    \gamma(\widehat{\vw};\vw)
    =\left\|(\widehat{\vw}-\vw)^\top\mX_\ell\right\|_2^2
    =(\widehat{\vw}-\vw)^\top\mC_\ell(\widehat{\vw}-\vw).
\]
Here, $\mC_\ell=\mX_\ell\mX_\ell^\top\in\R^{d_{\ell-1}\times d_{\ell-1}}$ is empirical second-order information associated with the activation input.
The OBS/GPTQ quantizer exploits the correlation among the entries of $\vw$ through $\mC_\ell$.
Suppose that the $i$-th entry of $\vw$ is quantized as $\widehat w_i=\mathcal Q_M(w_i)$, and denote its quantization displacement by
\(
    p_i=\widehat w_i-w_i.
\)
Instead of leaving all the remaining entries unchanged, one can compensate for $p_i$ by modifying the other unquantized entries of $\vw$.
Let $\vdelta_i$ denote the total change applied to $\vw$.
The optimal compensation is obtained by solving
\[
\begin{aligned}
    \underset{\vdelta_i}{\operatorname{minimize}}\quad&
    \vdelta_i^\top\mC_\ell\vdelta_i,\\
    \text{subject to}\quad&\ve_i^\top\vdelta_i=p_i,
\end{aligned}
\]
where $\ve_i$ is the $i$-th standard basis vector.
Assuming $\mC_\ell\succ\bm 0$, the solution is
\[
    \vdelta_i^\star
    =\textstyle \frac{p_i}{[\mC_\ell^{-1}]_{ii}}\mC_\ell^{-1}\ve_i.
\]
In the first step, the corresponding full-vector compensation is
\[
    \vw\leftarrow\vw+
    \textstyle\frac{\mathcal Q_M(w_i)-w_i}{[\mC_\ell^{-1}]_{ii}}
    \mC_\ell^{-1}\ve_i.
\]
The $i$-th entry of the updated vector is exactly $\mathcal Q_M(w_i)$.
In sequential quantization, the same compensation principle is applied to the remaining unquantized coordinates using the corresponding restricted curvature.
Applying this procedure sequentially to the entries of $\vw$ gives an OBS-style quantizer.
GPTQ uses the same second-order compensation principle, but organizes the sequential updates efficiently so that the rows of $\mW_\ell$ can be processed in parallel and the weight coordinates can be quantized in blocks.

\paragraph{Transform-Based Quantization.}
The OBS/GPTQ quantization schemes adjust the pre-quantized weight matrix using information from the covariance $\mC_\ell$.
When the covariance is poorly conditioned, some coordinate directions have a larger variance than others.
In this case, the uniform quantizer wastes resolution on low-variance directions while being too coarse for high-variance ones.
An orthogonal transformation applied before quantization can redistribute the variance more evenly across coordinates.

Let $\mR\in\R^{d\times d}$ be an orthogonal matrix satisfying $\mR^\top\mR=\mI$.
Recall the convention $\gamma=\|(\widehat{\vw}-\vw)^\top\mX_\ell\|_2^2$.
The layer output for a single weight row satisfies
\[
    \vw^\top\mX_\ell=(\mR\vw)^\top(\mR\mX_\ell).
\]
Defining $\widetilde{\mX}_\ell=\mR\mX_\ell$ and $\widetilde{\vw}=\mR\vw$ preserves the output exactly.
The rotated curvature is
\[
    \widetilde{\mC}_\ell
    =\widetilde{\mX}_\ell\widetilde{\mX}_\ell^\top
    =\mR\mC_\ell\mR^\top,
\]
and the OBS/GPTQ compensation rule may be applied in the rotated domain with $\widetilde{\mC}_\ell^{-1}$ in place of $\mC_\ell^{-1}$.
For the full weight matrix, the rotated weights are $\widetilde{\mW}_\ell=\mW_\ell\mR^\top$.

The goal of $\mR$ is to make the entries of $\widetilde{\mW}_\ell$ and $\widetilde{\mX}_\ell$ more uniformly distributed across coordinates, so that a fixed quantization step size introduces approximately equal distortion in every direction.
A random orthogonal $\mR$ can spread concentrated directions across the coordinate basis, while structured choices such as the randomized Hadamard transform provide a similar effect at lower computational cost~\citep{tseng2024quipsharp}.

\paragraph{Error Propagation.}
The layer-wise reconstruction objective measures the effect of a
weight error at the output of the quantized layer. An error introduced
at layer $\ell$ also passes through subsequent layers. 
We study its effect on the final network output.

For a calibration input $\vx_t$, let $\vx_{\ell,t}$ be the input to
layer $\ell$ and $\vz_{\ell,t}=\mW_\ell\vx_{\ell,t}$ its preactivation.
Write $\widehat{\mW}_\ell=\mW_\ell+\mQ_\ell$, where $\mQ_\ell$ is the
resulting weight-error matrix. Suppose that only layer $\ell$ is
perturbed. The preceding layers are unchanged, so $\vx_{\ell,t}$
is fixed, and the preactivation changes exactly by
\[
    \Delta\vz_{\ell,t}=\mQ_\ell\vx_{\ell,t}.
\]

Let $\vf_t=\mathcal F_{\{\mW_k\}_{k=1}^{L}}(\vx_t)$ be the
full-precision network output. Denote by $h_\ell$ the map from the
preactivation at layer $\ell$ to the final output, with all
subsequent weights fixed. Thus $\vf_t=h_\ell(\vz_{\ell,t})$.
Its Jacobian at the full-precision preactivation is
\[
    \mJ_{\ell,t}
    =\textstyle 
    \frac{\partial h_\ell(\vz)}{\partial\vz^\top}
    \Big|_{\vz=\vz_{\ell,t}}
    \in\R^{d_L\times d_\ell}.
\]
A first-order expansion gives
\[
    \widehat{\vf}_t-\vf_t
    \approx
    \mJ_{\ell,t}\mQ_\ell\vx_{\ell,t}.
\]
The preactivation change is exact under this single-layer
perturbation, and the downstream effect is linearized.
Averaging the predicted output changes over the
calibration set gives
\begin{equation}
\label{eq:jac-distortion}
\begin{aligned}
    D_{\mathrm{Jac},\ell}(\mQ_\ell)
    &:=\textstyle
    \frac{1}{N}\sum_{t=1}^{N}
    \|\mJ_{\ell,t}\mQ_\ell\vx_{\ell,t}\|_2^2
    =
    \operatorname{vec}(\mQ_\ell)^\top
    \mH_{\mathrm{Jac},\ell}^{\rm full}
    \operatorname{vec}(\mQ_\ell),
\end{aligned}
\end{equation}
where, using a column-wise vectorization, we have
\begin{equation}
\label{eqn:Hfull}
    \mH_{\mathrm{Jac},\ell}^{\rm full}
    =
    \textstyle \frac{1}{N}\sum_{t=1}^{N}
    \bigl(
        \vx_{\ell,t}\vx_{\ell,t}^{\top}
        \otimes
        \mJ_{\ell,t}^{\top}\mJ_{\ell,t}
    \bigr).
\end{equation}
Here, $D_{\mathrm{Jac},\ell}$ measures the network-output
distortion, and $\mH_{\mathrm{Jac},\ell}^{\rm full}$ is the
quadratic matrix on the vectorized weight error. 

The above characterization tracks an error 
introduced at layer $\ell$
through the subsequent network. 
It differs from the
upstream-error compensation studied by \cite{arai2025qep},
which accounts for errors from previously quantized layers
when reconstructing the current layer's output.

\section{Flexible Noise Shaping by a Causal Feedback Quantizer}
\label{sec:noise-shaping}
In the above derivations, we have seen that the Jacobian-weighted distortion is a function of the quantization error.
Directly optimizing this distortion is difficult because the instantaneous quantization errors depend on the inputs to the fixed uniform quantizer $\mathcal Q_M(\cdot)$.
In this section, we describe a causal feedback quantizer that uses error-feedback coefficients as design variables.

For convenience, we again consider the quantization of a single row of the weight matrix.
We drop the $t$ and $\ell$ indices momentarily.
Given a frozen network weight $\vw\in\R^d$, consider the preactivation output
\( \vy=\mX^\top\widehat{\vw}\),
where
\(\mX\in\R^{d\times N}\).
A causal feedback quantizer generates the quantized weights according to
\begin{subequations}
\label{eqn:efq}
\begin{align}
    \label{eqn:efq-b_n}
    b_i &= \textstyle w_i+\sum_{j<i}g_{ij}q_j,\\
    \label{eqn:efq-hatw}
    \widehat w_i &= \mathcal Q_M(b_i) = b_i + q_i,\\
    \label{eqn:efq-qn}
    q_i &= \widehat w_i-b_i,
\end{align}
\end{subequations}
where $g_{ij}\in\R$ are the feedback coefficients, 
$b_i$ is the instantaneous quantizer input, 
and $q_i$ is the instantaneous quantization error introduced when quantizing $b_i$. 
In the first equation, the summation is taken over $j<i$, that is, the errors introduced by quantizing prior indices.
Coupling $i = 1, \dots, d$, the error-feedback quantizer has the input-output relationship
\begin{equation}
    \widehat{\vw}=\vw+\mG\vq,
\end{equation}
where $\mG$ is a lower-triangular matrix with unit diagonal entries.
Thus, $\vq$ collects the instantaneous quantization errors, whereas $\mG\vq$ is the shaped quantization error.

Such error-feedback quantizers have historically been deployed in analog-to-digital and digital-to-analog converters (ADCs/DACs), specifically in the form of Sigma-Delta ($\Sigma\Delta$) modulation \citep{aziz1996overview_SigmaDelta}. 
In that context, the goal is to shape quantization noise toward a high-frequency range so that the baseband signal has little distortion.
In addition, $\Sigma\Delta$ modulation has been deployed in image halftoning in the form of error diffusion \citep{kite1997halftoning} and has recently been revisited for low-bit signal transmission in wireless communication \citep{shao2019onebitSigmaDelta, 10465600}.
In that context, the feedback coefficients can be adjusted to shape quantization noise away from target users.
In this work, we pursue noise shaping in model quantization.

Looking at \eqref{eqn:efq-qn}, it is difficult to characterize $q_i$ tractably without knowing the quantizer input $b_i$.
In general, $q_i$ depends on both $\mG$ and $\vw$, making the direct design of $\mG$ difficult.
We therefore make the following design assumption.
\begin{assumption}
\label{ass:white-noise}
The vector $\vq$ of quantization errors is random with second-order moment $\E\{\vq\vq^\top\}=\sigma_q^2\mI$,
independent of any other variables.
\end{assumption}
Consider the layer-wise distortion
\[
    \gamma(\mG)
    =\|\mX^\top(\widehat{\vw}-\vw)\|_2^2
    =\|\mX^\top\mG\vq\|_2^2
    =\vq^\top\mG^\top\mC\mG\vq,
\]
where $\mC=\mX\mX^\top$.
Under Assumption~\ref{ass:white-noise}, the expected loss is
\(
    \E\{\gamma(\mG)\}
    =\sigma_q^2\tr(\mG^\top\mC\mG).
\)
When treated as fixed, it does not affect the optimal design of $\mG$.
Hence, the design problem is posed as
\begin{equation}
\label{eqn:design-of-G}
\begin{array}{rl}
    \underset{\mG}{\operatorname{minimize}}&\tr(\mG^\top\mC\mG)\\
    \text{subject to}&g_{ij}=0,\quad j>i,\qquad g_{ii}=1.
\end{array}
\end{equation}
Assuming $\mC\succ\bm0$, possibly after diagonal regularization, \eqref{eqn:design-of-G} is strictly convex.
One can leverage the KKT conditions to obtain the global minimizer of \eqref{eqn:design-of-G}.
Consider the Lagrangian
\begin{equation}
\label{eqn:Lagrangian-of-G}
    \mathcal L=\tr(\mG^\top\mC\mG)
    +\tr\bigl(\bm\Lambda^\top(\mG-\mI)\bigr),
\end{equation}
where $\bm\Lambda$ is an upper-triangular matrix collecting the Lagrange multipliers associated with the constraints on $g_{ij}$.
The stationarity condition reads
\[
    \nabla_{\mG}\mathcal L=2\mC\mG+\bm\Lambda=\bm0,
\]
implying $\mC\mG=-\frac12\bm\Lambda$.
We see that $\mC\mG$ is upper triangular.
This implies that $\mG^\top(\mC\mG)$ is also upper triangular.
Moreover, $\mG^\top\mC\mG$ is symmetric.
Combining these facts yields $\mG^\top\mC\mG:=\mD$, where $\mD$ is diagonal.
Taking inverse on both sides gives
\[
    (\mG^\top\mC\mG)^{-1}=\mD^{-1}
    \quad\Longrightarrow\quad
    \mC^{-1}=\mG\mD^{-1}\mG^\top.
\]
Consider the symmetric LDL decomposition $\mC^{-1}=\mL\bar{\mD}\mL^\top$, where $\mL$ is unit lower triangular and $\bar{\mD}$ is diagonal.
Comparing these equations, the optimizer of \eqref{eqn:design-of-G} is $\mG^\star=\mL$.
This resembles the fixed-order covariance-based feedback structure associated with GPTQ/LDLQ \citep{chee2023quip, birnick2025iclr_babai, chen2026babai}.

The above description resembles the GPTQ-style feedback structure from the perspective of designing the coefficients of a causal error-feedback quantizer.
This permits us to generalize the design criterion from layer-wise distortion
to a general loss function $f$, i.e., 
\begin{equation}
\label{eqn:design-of-G-general}
\begin{array}{rl}
    \underset{\mG}{\operatorname{minimize}}&f(\mG;\bm\Theta)\\
    \text{subject to}&g_{ij}=0,\quad j>i,\qquad g_{ii}=1,
\end{array}
\end{equation}
where $\bm\Theta$ denotes network information that may extend beyond the data covariance $\mC$.

{\color{black} 
The network information $\bm \Theta$ may include outputs of
already-quantized upstream layers. Consequently,
$f(\mG;\Theta)$ can also express
mismatched-input reconstruction when those outputs
are held fixed during layerwise design; see
Appendix~\ref{app:related-objectives}.
}

As a design example, let $\bar{\mC}_\ell=\mC_\ell/N$ denote the sample-averaged input covariance.
For weight row $r$, define the input-coordinate Jacobian metric
\begin{equation}
\label{eqn:Hrow}
\mH_{{\rm Jac},\ell,r}^{\rm row}
    :=\textstyle \frac1N\sum_{t=1}^N
      \|\mJ_{\ell,t}\ve_r\|_2^2
      \vx_{\ell,t}\vx_{\ell,t}^\top
    \, \in \, \R^{d_{\ell-1}\times d_{\ell-1}},
\end{equation}
where $\ve_r\in\R^{d_\ell}$ selects the $r$-th output coordinate.
A row-wise feedback design may then use
\begin{equation}
\label{eqn:fHrow}
    f_r(\mG;\bm\Theta)
    =\tr\!\left[\mG^\top
      \bigl(\bar{\mC}_\ell
        +\alpha s_\ell\mH_{{\rm Jac},\ell,r}^{\rm row}\bigr)
      \mG\right],
\end{equation}
where $\alpha\geq0$ controls the tradeoff and $s_\ell>0$ is a fixed layer-wise scale.
When a feedback matrix is shared across rows, 
we use the metric
\begin{align}
\label{eqn:Hiso}
\mH_{{\rm Jac},\ell}^{\rm iso}
   &:= \frac1{d_\ell}\sum_{r=1}^{d_\ell}
       \mH_{{\rm Jac},\ell,r}^{\rm row} 
    = \frac1N \sum_{t=1}^N
       \frac{\|\mJ_{\ell,t}\|_F^2}{d_\ell}
       \vx_{\ell,t}\vx_{\ell,t}^\top.
\end{align}

\begin{table*}[t]
\centering
\caption{Test-set output distortion over $100$ independently
generated networks. The fixed-$\alpha$ columns report mean
distortion when the same $\alpha$ is used for every network;
$\alpha=0$ is the GPTQ-style baseline. In the final three
columns, $\alpha$ is selected separately for each network
using calibration data only. Reduction is relative to
$\alpha=0$, and Win/Tie/Loss counts networks on which the
selected method has lower, equal, or higher test distortion
than that baseline. Boldface in the fixed-$\alpha$ columns
marks the lowest mean in each row.}
\label{tab:synthetic-alpha-combined}
\scriptsize
\setlength{\tabcolsep}{3pt}
\resizebox{.9\textwidth}{!}{%
\begin{tabular}{c|ccccccccc|ccc}
\toprule
& \multicolumn{9}{c|}{Fixed $\alpha$: mean test distortion}
& \multicolumn{3}{c}{Per-network $\alpha$ selection}\\
\cmidrule(lr){2-10}\cmidrule(lr){11-13}
$M$
& $0$ & $0.01$ & $0.03$ & $0.1$ & $0.2$
& $0.4$ & $0.6$ & $0.8$ & $0.99$
& Mean distortion & Reduction & Win/Tie/Loss\\
\midrule
2
& 1.985016 & \textbf{1.978903} & 2.024264
& 2.032677 & 2.045072 & 2.038097
& 2.036754 & 2.032965 & 2.046547
& 1.691480 & 14.79\% & 75/20/5\\
4
& 0.136493 & 0.137417 & 0.133098
& 0.133740 & 0.134332 & 0.133041
& 0.131872 & \textbf{0.130650} & 0.131327
& 0.112534 & 17.55\% & 75/21/4\\
8
& 0.022783 & 0.022621 & 0.022686
& 0.022051 & \textbf{0.021853} & 0.022372
& 0.022070 & 0.021908 & 0.022043
& 0.018986 & 16.67\% & 72/24/4\\
12
& 0.008864 & 0.008892 & 0.008792
& \textbf{0.008754} & 0.008883 & 0.008964
& 0.009062 & 0.008925 & 0.008988
& 0.007633 & 13.88\% & 73/20/7\\
16
& 0.004611 & 0.004536 & \textbf{0.004499}
& 0.004568 & 0.004534 & 0.004555
& 0.004587 & 0.004654 & 0.004631
& 0.003855 & 16.39\% & 72/27/1\\
\bottomrule
\end{tabular}%
}
\end{table*}

\paragraph{Numerical Example.}
We investigate whether the flexibility provided by $\mG$ in \eqref{eqn:design-of-G-general} brings a benefit to network quantization.
We construct a five-layer fully connected network with widths $12\to16\to16\to12\to8\to3$.
The first four layers are followed by the $\tanh$ activation, and the output layer is linear.
The full-precision weights are drawn independently from zero-mean Gaussian distributions, with variances scaled according to the layer input dimensions.
The network input data are generated from a four-dimensional latent Gaussian variable followed by a fixed linear mixing matrix and additive Gaussian noise.
We draw $256$ samples for calibration and an independent set of $1024$ samples for evaluation.
All covariance and Jacobian matrices are computed using the full-precision network on the calibration set.

We consider $M\in\{2,4,8,12,16\}$ levels of uniform quantization.
For each weight row, the quantization alphabet is fixed before applying error feedback and shared by all choices of $\alpha$.
For each $M$, we sweep over $\alpha\in\{0,0.01,0.03,0.1,0.2,0.4,0.6,0.8,0.99\}$.
The row-wise Jacobian term is scaled by a common layer-wise factor $s_\ell$ before applying $\alpha$.
The output distortion is measured by the test MSE between the quantized and full-precision networks.
The experiment is repeated over $100$ independently generated networks and datasets.

The numerical results are presented in Table~\ref{tab:synthetic-alpha-combined}.
When $\alpha$ is selected using calibration data, the Jacobian-aware design reduces the mean test distortion relative to the GPTQ-style baseline for each $M$, with relative reductions ranging from $13.88\%$ to $17.55\%$.
It also achieves lower test distortion on at least $72\%$ of the random instances for every $M$.
The complete sweep shows that the preferred value of $\alpha$ depends on the quantization resolution and that a fixed positive $\alpha$ does not necessarily improve upon $\alpha=0$.
These results indicate a potential benefit from a flexibly designed feedback matrix.
The tradeoff factor $\alpha$ provides a degree of freedom to balance local layer-wise distortion and downstream error propagation.

\section{Joint Design of Rotation and Error Feedback}
\label{sec:joint_rotation_feedback}
Section~\ref{sec:background} described how an orthogonal rotation can redistribute weight outliers before quantization.
Here, we consider its distinct role in shaping quantization error jointly with the feedback matrix.
Following the convention $\widetilde{\vw}=\mR\vw$ and $\widehat{\widetilde{\vw}}-\widetilde{\vw}=\mG\vq$, the error in the original coordinates is $\widehat{\vw}-\vw=\mR^\top\mG\vq$.
For a fixed positive-definite effective metric $\mH$, the resulting trace problem is
\begin{equation}
\label{eq:joint-trace-design}
    \min_{\substack{\mR^\top\mR=\mI\\\mG\ \text{unit lower triangular}}}
    f(\mG,\mR;\mH)
    :=\tr\!\left(\mG^\top\mR\mH\mR^\top\mG\right).
\end{equation}
The metric may, for example, be the local covariance $\mC_\ell$ or a specified Jacobian-aware input-coordinate metric.
Rotation alone leaves the trace unchanged: $f(\mI,\mR;\mH)=\tr(\mH)$.
Its effect in \eqref{eq:joint-trace-design} arises from changing the coordinates in which triangular feedback operates.

For a fixed $\mR$, the reverse LDL factorization is
\[
    \mR\mH\mR^\top=\mU_{\mR}\mD_{\mR}\mU_{\mR}^\top,
\]
where $\mU_{\mR}$ is unit upper triangular and $\mD_{\mR}$ is diagonal.
The feedback solution from Section~\ref{sec:noise-shaping} then gives
\[
    \mG_{\mR}^\star=\mU_{\mR}^{-\top},\qquad
    \min_{\mG}f(\mG,\mR;\mH)=\tr(\mD_{\mR}).
\]
Since $\det(\mD_{\mR})=\det(\mH)$, the arithmetic-geometric mean inequality yields the lower bound $n\det(\mH)^{1/n}$.
\vspace{-3mm}
\begin{proposition}[Joint rotation-feedback optimum]
\label{prop:joint_rg_global_optimum}
Let $\mH\succ\bm0$ and $\eta=\det(\mH)^{1/n}$.
Then
\[
    \min_{\substack{\mR^\top\mR=\mI\\\mG\ \text{unit lower triangular}}}
    f(\mG,\mR;\mH)=n\eta.
\]
To construct an optimizer, take a geometric mean decomposition $\mH^{1/2}=\mQ\mT\mV^\top$, where $\mQ,\mV$ are orthogonal and $\mT$ is upper triangular with diagonal entries $\sqrt\eta$.
Then
\[
    \mR^\star=\mQ^\top,\qquad
    \mG^\star=\sqrt\eta\,\mT^{-\top}.
\]
\end{proposition}
Indeed, $\mR^\star\mH(\mR^\star)^\top=\mT\mT^\top$, and substitution gives $(\mG^\star)^\top\mR^\star\mH(\mR^\star)^\top\mG^\star=\eta\mI$, attaining the lower bound.
Thus, rotation equalizes the feedback-adjusted sensitivities, while feedback removes their triangular coupling.
Replacing $\mH$ by $\mH+\lambda\mI\succ\bm0$ gives the corresponding damped design.
The optimum concerns a fixed metric and fixed second moment of the instantaneous quantization errors.
In the experiments below, the construction is applied independently to input-coordinate blocks, so its attained bound is blockwise rather than an unrestricted full-layer optimum.


\subsection{Finite-level quantization: the role of scale}
\label{sec:finite-level-scale}

Proposition~1 minimizes $f(\mG,\mR;\mH)$ for a fixed metric and a common,
fixed second moment of the instantaneous errors. At finite resolution,
the error statistics depend on the quantization scale and the chosen
rotation and feedback. For the min--max grids in our MLP and GPT-2
experiments, a rotated weight row or block has step size
\[
\Delta(\mR)=\frac{\max_i(\mR\vw)_i-\min_i(\mR\vw)_i}{M-1}.
\]
Although fixed during feedback, this scale need not be shared across
rotations. Under the idealized approximation of independent, uniform
rounding errors on $[-\Delta/2,\Delta/2]$ without overload,
\[
\mathbb E[D]\approx\frac{\Delta(\mR)^2}{12}\,f(\mG,\mR;\mH),
\]
where $D=(\widehat \vw-\vw)^\top \mH(\widehat \vw-\vw)$.
Thus, a smaller trace can be offset by a larger scale.
Orthogonality preserves the weight norm, not its coordinate range,
and the trace construction does not optimize this range.

Feedback can also move quantizer inputs outside the fitted interval,
producing saturation and nonisotropic errors. Widening the interval to
accommodate these inputs increases $\Delta$ at fixed $M$, trading
resolution for range. These effects explain why the trace optimum is
not a finite-level performance guarantee. Appendix~C verifies trace
attainment without consistent finite-level gains; Table~3 likewise
shows that RHT can outperform joint design. These results do not
isolate scale as the sole cause. The proposition is therefore a
fixed-statistics noise-shaping benchmark, not an optimum over scale,
center, and finite-grid errors.


\section{Numerical Results}
\label{sec:numerical_results}

\subsection{PTQ of Trained MLPs}
\label{sec:mlp_results}

We study weight-only post-training quantization of
five-linear-layer tanh networks trained on MNIST and CIFAR-10 \citep{lecun1998mnist, krizhevsky2009learning}.
The MNIST network has widths
$784\to128\to128\to64\to32\to10$, and the CIFAR-10
network has widths $3072\to128\to128\to64\to32\to10$.
Images are flattened before entering the networks, and tanh
activations follow the first four linear layers.
We compare round-to-nearest quantization (RTN),
covariance-based causal error feedback (EFQ), randomized
Hadamard transforms (RHT) followed by feedback,
objective-aware (based on equations \ref{eqn:fHrow} and \ref{eqn:Hrow} or \ref{eqn:Hiso}), and joint rotation--feedback design (based on Proposition \ref{prop:joint_rg_global_optimum}).

All methods use fixed quantization grids with
$M\in\{2,4,8,16\}$ levels. Input coordinates within each
weight row are partitioned into consecutive blocks of at
most $32$, with rotation and feedback designed independently
within each block. For objective-aware methods, $\alpha$ is selected among candidates in $\{0,0.03,0.1,0.3,1\}$ using data
disjoint from metric calibration and testing. 
The blockwise design metric combines 
the sample-averaged input covariance
with a trace-normalized, 
row-isotropic Jacobian metric (\eqref{eqn:Hiso}) and
diagonal damping ($\lambda$$>$0). Candidate selection uses the full
first-order network-output distortion (equations \ref{eq:jac-distortion} and \ref{eqn:Hfull}) , whereas construction of each feedback and rotation design uses the row-isotropic
trace surrogate (based on the diagonalization of \eqref{eqn:Hiso}). Further details appear in
Appendix~\ref{app:mlp-protocol}.
Table~\ref{tab:mlp_joint_results} reports test logit MSE relative
to the full-precision model and accuracy drop in percentage
points.

\begin{table*}[t]
\centering
\caption{Post-training quantization of MNIST- and
CIFAR-10-trained MLPs. Each entry reports test logit MSE /
accuracy drop in percentage points, averaged over 100 trained
checkpoints. Mean full-precision accuracy is $97.0539\%$ on
MNIST and $47.5550\%$ on CIFAR-10. Lower is better for both
metrics. Boldface marks the lowest unrounded mean within each
dataset and $M$.}
\label{tab:mlp_joint_results}
\footnotesize
\resizebox{0.95\textwidth}{!}{%
\begin{tabular}{llcccc}
\toprule
Dataset & Method & $M=2$ & $M=4$ & $M=8$ & $M=16$\\
\midrule
\multirow{6}{*}{MNIST}
 & RTN
 & $8.606688/32.57$ & $0.382275/0.92$
 & $0.056979/0.11$ & $0.011988/0.03$\\
 & Local EFQ
 & $1.695767/5.12$ & $0.084901/\mathbf{0.20}$
 & $0.013702/0.04$ & $0.002940/\mathbf{-0.0008}$\\
 & Local EFQ + RHT
 & $3.921692/15.91$ & $0.139418/0.27$
 & $0.023697/0.04$ & $0.004970/0.01$\\
 & Obj. EFQ
 & $\mathbf{1.637242}/\mathbf{4.83}$
 & $\mathbf{0.082757}/0.21$
 & $\mathbf{0.013351}/\mathbf{0.03}$
 & $\mathbf{0.002856}/-0.0003$\\
 & Obj. EFQ + RHT
 & $3.839215/15.08$ & $0.135396/0.25$
 & $0.023065/0.04$ & $0.004850/0.01$\\
 & Obj. joint $\mR,\mG$
 & $8.672428/44.34$ & $0.140144/0.49$
 & $0.014719/0.03$ & $0.002961/0.00$\\
\midrule
\multirow{6}{*}{CIFAR-10}
 & RTN
 & $5.123931/14.39$ & $0.260287/2.13$
 & $0.046357/0.43$ & $0.010024/0.10$\\
 & Local EFQ
 & $1.838424/13.18$ & $0.105956/\mathbf{0.93}$
 & $0.018626/\mathbf{0.14}$ & $0.004025/0.04$\\
 & Local EFQ + RHT
 & $2.955296/18.37$ & $0.139829/1.24$
 & $0.024744/0.21$ & $0.005300/0.06$\\
 & Obj. EFQ
 & $\mathbf{1.815878}/\mathbf{13.05}$
 & $\mathbf{0.104643}/0.93$
 & $\mathbf{0.018366}/0.14$
 & $\mathbf{0.003951}/\mathbf{0.04}$\\
 & Obj. EFQ + RHT
 & $2.896634/18.06$ & $0.138125/1.21$
 & $0.024415/0.22$ & $0.005216/0.05$\\
 & Obj. joint $\mR,\mG$
 & $3.448898/19.50$ & $0.122654/1.07$
 & $0.020478/0.22$ & $0.004366/0.04$\\
\bottomrule
\end{tabular}%
}
\end{table*}

Objective-aware EFQ has the lowest mean test logit MSE
at every tested resolution on both datasets, although
its advantage over covariance-based EFQ narrows at
larger $M$. Accuracy drop does not always follow the
same ordering: covariance-based EFQ has a slightly
lower mean accuracy drop at $M=4$ on both datasets,
despite its higher logit MSE. Joint rotation-feedback
performs poorly at $M=2$ on MNIST. A separate paired
diagnostic in the appendix shows that joint design
attains its blockwise fixed-statistics trace bound
without necessarily reducing realized finite-level
error. That diagnostic fixes $\alpha=0$ and is distinct
from the selected-$\alpha$ comparisons reported here.

\begin{table*}[t]
\centering
\caption{Post-training quantization on five language models.
Each entry reports PPL / next-token accuracy (\%) /
final-logit MSE. Lower PPL and MSE and higher accuracy
are better. Boldface marks the best of the six methods
for each model and bit width. ``Local'' and
``objective-aware'' specify the design metric; EFQ,
RHT+EFQ, and joint $\mR,\mG$ specify the construction.
}
\label{tab:llm_combined_main}
\resizebox{\textwidth}{!}{
 \fontsize{7}{8.2}\selectfont
\setlength{\tabcolsep}{0.5pt}
\renewcommand{\arraystretch}{1.12}
\spaceskip=0.20em plus 0.03em minus 0.03em
\begin{tabular}{@{}llcccccc@{}}
\toprule
Model & $M$
& Local EFQ
& Local EFQ+RHT
& Local joint $\mR,\mG$
& Obj.-EFQ
& Obj.-EFQ+RHT
& Obj. joint $\mR,\mG$ \\
\midrule
GPT-2
& 8
& 118.54 / 24.21 / 257.89
& 105.41 / 24.60 / 221.42
& 112.66 / 24.26 / 186.16
& 119.03 / 24.45 / 222.64
& \textbf{105.38} / 24.40 / 206.28
& 106.94 / \textbf{24.65} / \textbf{182.59} \\
& 16
& 99.65 / 25.50 / 85.48
& 100.69 / 25.30 / 34.73
& 98.41 / \textbf{26.14} / 31.10
& 99.60 / 25.40 / 72.64
& 99.95 / 25.50 / \textbf{28.13}
& \textbf{98.25} / 26.09 / 57.73 \\
\midrule
Llama-2-7B
& 8
& 11.312 / 50.22 / 3.989
& 9.120 / 53.15 / 2.077
& 9.893 / 51.70 / 2.794
& 11.002 / 50.81 / 3.745
& \textbf{8.704 / 53.73 / 1.739}
& 9.338 / 52.55 / 2.253 \\
& 16
& 7.808 / 55.69 / 0.791
& 7.450 / 56.34 / 0.347
& 7.563 / 56.06 / 0.498
& 7.736 / 55.84 / 0.685
& \textbf{7.431 / 56.44 / 0.309}
& 7.521 / 56.18 / 0.453 \\
\midrule
Llama-2-13B
& 8
& 8.209 / 54.56 / 0.842
& 7.585 / 55.73 / 0.619
& 8.238 / 54.52 / 0.806
& 7.929 / 55.13 / 0.765
& \textbf{7.501 / 55.97 / 0.579}
& 7.989 / 54.86 / 0.733 \\
& 16
& 6.714 / 58.00 / 0.201
& 6.611 / 58.25 / 0.137
& 6.671 / 58.03 / 0.182
& 6.630 / 58.15 / 0.168
& \textbf{6.593 / 58.27 / 0.126}
& 6.675 / 58.04 / 0.168 \\
\midrule
Qwen3-8B
& 8
& 20.227 / 43.52 / 5.933
& 16.485 / 46.91 / 2.583
& 16.857 / 46.45 / 3.367
& 20.531 / 43.76 / 4.616
& \textbf{16.118 / 47.08 / 2.536}
& 16.839 / 46.50 / 3.008 \\
& 16
& 14.140 / 49.53 / 1.156
& 13.469 / 50.23 / 0.570
& 13.473 / 50.12 / 0.680
& 13.750 / 49.71 / 0.952
& \textbf{13.312 / 50.32 / 0.554}
& 13.449 / 50.18 / 0.659 \\
\midrule
Qwen3-14B
& 8
& 16.487 / 46.47 / 2.774
& 13.924 / 49.27 / 1.709
& 14.196 / 48.94 / 1.992
& 16.561 / 46.63 / 2.773
& \textbf{13.734 / 49.44 / 1.658}
& 14.077 / 49.00 / 1.947 \\
& 16
& 12.447 / 51.35 / 0.750
& 11.984 / 51.88 / 0.407
& 12.067 / 51.76 / 0.482
& 12.329 / 51.37 / 0.716
& \textbf{11.912 / 51.93 / 0.385}
& 12.045 / 51.84 / 0.475 \\
\bottomrule
\end{tabular}
}
\end{table*}

\subsection{PTQ of Language Models}
\label{sec:llm_quantization}

We evaluate feedback design on GPT-2 \citep{radford2019language_gpt2},
Llama-2 \citep{touvron2023llama2}, 
and Qwen3 \citep{qwen3technicalreport} 
language models. 
Table~\ref{tab:llm_combined_main} reports
perplexity, next-token accuracy, and final-logit MSE at
$M\in\{8,16\}$ levels. ``Local'' denotes a projection-input
metric, while ``objective-aware'' denotes a
downstream-sensitive logit metric. Under each metric, we
compare EFQ without rotation, RHT followed by EFQ, and
joint rotation-feedback design.

For GPT-2, we quantize the $48$ attention and MLP
projections in input-coordinate blocks of size $32$,
retaining full-precision embeddings. Objective-aware
feedback uses stochastic downstream-sensitivity estimates;
$\alpha$ is selected by validation logit MSE. The reported
GPT-2 rows use $8064$ scored positions and a separate
evaluation protocol from the larger models. 
Its
full-precision perplexity and next-token accuracy are
$96.35$ and $26.09\%$, respectively.

For Llama-2-7B, Llama-2-13B, Qwen3-8B, and Qwen3-14B,
metric calibration uses seed $7$ and $128$
non-overlapping WikiText-2 \citep{merity2016pointer_wikitext2}
training sequences of length
$512$. Evaluation uses $340326$ scored positions for
each Llama model and $298424$ for each Qwen model.
Comparisons in the table are made within each model and
bit width, not across the two evaluation protocols.

On the four larger models, objective-aware design lowers
final-logit MSE and raises next-token accuracy in all
$24$ comparisons matched by model, bit width, and
construction; it lowers perplexity in $21$ comparisons. 
Objective-aware
RHT+EFQ has the best three reported metrics among the six
methods in each of settings considered.
GPT-2 is less uniform: at three and four bits, its best
perplexity, accuracy, and logit MSE do not all belong to
the same method.

On the four larger models, 
RHT+EFQ outperforms joint rotation-feedback 
on all three reported metrics in the 16 comparisons. 
This finite-level ordering does not contradict
joint design's blockwise trace optimum: 
the objective does not account for rotation-dependent finite-grid errors.

\subsection{Direct Feedback Tuning in QAT}
\label{sec:qat_results}

The PTQ experiments above choose a design metric and
compute the corresponding feedback matrix through 
the trace problem. We next
consider a complementary route: updating the free
entries of $\mG$ against the loss of the
hard-quantized network using a straight-through
backward rule. 
\vspace{-3mm}
\paragraph{Synthetic regression.}
We consider a two-linear-layer regression network
with dimensions $4\to6\to3$ and a $\tanh$ hidden
activation. One unit lower-triangular feedback
matrix is shared across the weight rows of each
layer. The data-loss objective is
\[
    \min_{\mW,\mG}\; \textstyle
    \frac{1}{N_{\rm tr}d_L}
    \sum_{t=1}^{N_{\rm tr}}
    \left\|
        {\cal F}_{\mathcal Q_M(\mW;\mG)}(\vx_t)
        -\vy_t
    \right\|_2^2.
\]
Here $\mathcal Q_M(\mW;\mG)$ applies the causal
quantizer in \eqref{eqn:efq} separately to each
weight row in input-coordinate order. The forward
pass uses hard nearest rounding on fixed per-row
grids; the backward pass uses a straight-through
surrogate derivative.

Across $100$ independently generated datasets,
all methods within a trial start from the same
fitted full-precision checkpoint. Each trial
has $160$ training, $80$ validation, and $256$
test pairs, with $M\in\{2,4,8\}$. Hard-quantized
validation MSE selects the checkpoint. Table~\ref{tab:tiny_g_qat}
reports test MSE against the generated targets;
Appendix~\ref{app:qat-details} gives the full
training protocol.

\begin{table}[t]
\centering
\caption{Mean test MSE over $100$ paired synthetic
regression trials; lower is better. Feedback-only
tuning holds the network fixed and starts from
covariance-designed feedback. Both displayed QAT
arms start from $\mG=\mI$. Boldface marks the
lowest mean among the quantized methods shown.}
\label{tab:tiny_g_qat}

\fontsize{7}{8.2}\selectfont
\setlength{\tabcolsep}{2.5pt}
\renewcommand{\arraystretch}{1.12}

\begin{tabular}{@{}lccc@{}}
\toprule
Method & $M=2$ & $M=4$ & $M=8$\\
\midrule
Full precision
 & 0.01391 & 0.01391 & 0.01391\\
RTN
 & 0.63739 & 0.06576 & 0.02395\\
\midrule
Local PTQ
 & 0.48603 & 0.05473 & 0.02078\\
Feedback QAT(G)
 & 0.28594 & 0.04055 & 0.01899\\
\midrule
Weight-only QAT; $\mG=\mI$ fixed
 & 0.25458 & 0.04118 & 0.01914\\
Joint QAT; $\mG_0=\mI$
 & \textbf{0.24524} & \textbf{0.04013}
 & \textbf{0.01784}\\
\bottomrule
\end{tabular}
\end{table}

Tuning $\mG$ with the fitted network fixed lowers
mean test MSE relative to covariance-feedback PTQ
at each tested resolution. In the matched
identity-initialized QAT comparison, joint
training gives lower mean test MSE than
weight-only QAT at all three resolutions.
The complete comparison in
Appendix~\ref{app:qat-details} also includes
covariance-initialized QAT arms.

\paragraph{Small character-level language model.}
We also evaluate feedback tuning on a character-level
nanoGPT model trained on the Shakespeare character
dataset \citep{karpathy2022nanoGPT}. 
It has 2 transformer blocks, 2 attention
heads, and embedding width $64$. 
We quantize 4
attention and MLP projection matrices per block in
input-coordinate blocks of size $32$.

For each of ten seeds, all arms start from the same
full-precision checkpoint trained for $200$ steps.
We compare RTN, covariance-feedback PTQ,
feedback-only tuning with network coefficients
fixed, weight-only QAT with $\mG=\mI$, and joint
QAT initialized at $\mG=\mI$. The tuning arms
receive $100$ additional steps. An unquantized
copy of the checkpoint also receives $100$
additional steps as a training-budget reference.

\begin{table}[t]
\centering
\caption{Held-out next-character perplexity on the Shakespeare
character dataset, averaged over ten paired seeds. Perplexity is
computed for each seed before averaging; lower is better.
Evaluation uses the held-out second half of the validation file.
Boldface marks the lowest mean among quantized methods.}
\label{tab:nanogpt-feedback-qat}

\fontsize{7}{8.2}\selectfont
\setlength{\tabcolsep}{2.2pt}
\renewcommand{\arraystretch}{1.12}

\begin{tabular}{@{}lcccc@{}}
\toprule
Method & $M=2$ & $M=4$ & $M=8$ & $M=16$\\
\midrule
Init. Full precision
 & 12.967 & 12.967 & 12.967 & 12.967\\
Cont. Full precision
 & 12.374 & 12.374 & 12.374 & 12.374\\
\midrule
RTN
 & 18.282 & 13.046 & 12.986 & 12.970\\
Local PTQ
 & 13.629 & 12.999 & 12.969 & 12.970\\
Feedback QAT ($\mG$)
 & 13.074 & 12.842 & 12.865 & 12.910\\
Weight QAT ($\mW$)
 & 12.561 & 12.404 & 12.390 & 12.382\\
Joint QAT ($\mW, \mG$)
 & \textbf{12.528} & \textbf{12.392}
 & \textbf{12.381} & \textbf{12.380}\\
\bottomrule
\end{tabular}
\end{table}

Feedback-only tuning has lower mean held-out
perplexity than fixed covariance feedback at
each tested level. Joint QAT has a slightly
lower unrounded mean than matched weight-only
QAT, but the differences are small; at $M=16$,
both means round to $12.38$. Continued
full-precision training has lower mean
perplexity than either QAT arm at every level.
This experiment demonstrates
that flexible feedback coefficients
can be trainable on language models.

\section{Conclusions} 
This work formulated error feedback 
as a quantizer-design problem.
We treated error feedback coefficients as design variables
rather than a prescribed compensation rule. 
For a specified input-coordinate metric 
and a fixed isotropic second moment of the instantaneous quantization errors,
this viewpoint yields covariance-based feedback
as a reference design
and a constructive optimum for joint design on rotation and feedback. 
Experiments show that adapting the feedback criterion can improve output fidelity, and feedback coefficients can also be 
jointly trained with model weights in a small QAT setting. 
They expose a limitation of the trace formulation: 
attaining its optimum does not necessarily 
reduce realized error under finite-level quantization. 
The resulting framework offers a way to 
formulate and test feedback designs 
beyond a single reconstruction objective. 
A key direction for future work 
is to account for how feedback and
rotation change the errors 
produced by the quantizer, 
and to examine whether learned feedback remains 
useful in larger pretrained models. 
In this sense, the contribution 
is both a characterization of what fixed-statistics design 
can achieve 
and 
a starting point for quantizer designs that 
reflect realized network behavior more closely.

\bibliography{references}
\appendix

\section{Related Compensation Objectives}
\label{app:related-objectives}

The abstraction $f(\mG_\ell;\Theta_\ell)$ separates the
feedback matrix being designed from the context held fixed
during that design. Let
$\widehat{\mW}_\ell(\mG_\ell)$ be the weights produced by
our causal quantizer and define
\[
\mE_\ell(\mG_\ell)
=
\widehat{\mW}_\ell(\mG_\ell)-\mW_\ell.
\]
For a layerwise design problem, $\Theta_\ell$ may contain
the original weights, quantization grid, calibration data,
design metric, and the outputs of already-quantized
upstream layers. In particular, let
$\mX_\ell$ and $\widetilde{\mX}_\ell$ contain the
full-precision and upstream-quantized inputs, respectively,
as columns. Both are held fixed while designing
$\mG_\ell$.

This notation permits a mismatched-input reconstruction
criterion:
\[
\begin{aligned}
f_{\rm M}(\mG_\ell;\Theta_\ell)
&=
\left\|
\mW_\ell\mX_\ell-
\widehat{\mW}_\ell(\mG_\ell)
\widetilde{\mX}_\ell
\right\|_F^2
=
\left\|
\mW_\ell(\mX_\ell-\widetilde{\mX}_\ell)
-
\mE_\ell(\mG_\ell)\widetilde{\mX}_\ell
\right\|_F^2.
\end{aligned}
\]
The first term inside the second norm records the
upstream-input mismatch; the second records the effect
of the current layer's quantized weights. Thus,
upstream mismatch can be represented in $\Theta_\ell$
without changing the design variable in Eq.~(5).
This criterion is generally not the homogeneous
fixed-statistics trace objective solved by our
rotation-feedback proposition.

For the comparison below, let
$\ve_{\ell,r}(\mG_\ell)$ be row $r$ of
$\mE_\ell(\mG_\ell)$, written as a column vector.
To describe end-loss-gradient weighting, define
\[
\mF_{\ell,r}
=
\frac{1}{N}\sum_{t=1}^{N}
\left(
\frac{\partial\mathcal L_t}
     {\partial z_{\ell,r,t}}
\right)^2
\vx_{\ell,t}\vx_{\ell,t}^{\top}.
\]
This row-wise metric is distinct from the
output-Jacobian metric used in our experiments.

\begin{table*}[t]
\centering
\caption{Related compensation criteria viewed through
$f(\mG_\ell;\Theta_\ell)$. 
}
\label{tab:related-feedback-objectives}
\small
\renewcommand{\arraystretch}{1.9}  
\setlength{\tabcolsep}{4pt}
\begin{tabular}{@{}
    p{0.2\textwidth}
    p{0.35\textwidth}
    p{0.4\textwidth}
@{}}
\toprule
Method & Compensation criterion
       & Relationship to our feedback design\\
\midrule

GPTQ / QuIP-LDLQ
\citep{frantar2023gptq,chee2023quip}
&
Layer-output quadratic
$\tr(\mE_\ell\mC_\ell\mE_\ell^\top)$.
&
The closest linear-feedback precedent. With
the input covariance as metric, our
fixed-isotropic-error design surrogate is
proportional to
$\tr(\mG_\ell^\top\mC_\ell\mG_\ell)$.
\\\hline

GuidedQuant
\citep{kim2025guidedquant}
&
End-loss-gradient-weighted row quadratics,
$\sum_r
\ve_{\ell,r}^{\top}\mF_{\ell,r}\ve_{\ell,r}$.
&
The criterion can be evaluated on
$\ve_{\ell,r}(\mG_\ell)$. Its weighting
uses end-loss gradients rather than our
output-Jacobian metric; this objective-level
connection does not identify the two
quantization algorithms.
\\\hline

YAQA
\citep{tseng2026yaqa}
&
A structured quadratic approximation to
full-model output-distribution KL.
&
Its criterion can be restricted to
$\operatorname{vec}(\mE_\ell(\mG_\ell))$.
Its adaptive rounding also uses output-side
feedback, beyond a single shared
input-coordinate $\mG_\ell$.
\\\hline

QEP
\citep{arai2025qep}
&
Compensation for quantization errors
propagated from preceding layers.
&
Upstream-quantized states can be included
in $\Theta_\ell$, permitting an
upstream-error-aware objective for
$\mG_\ell$. This is a restriction to
our quantizer, not an equivalence with
QEP's procedure.
\\\hline

Qronos
\citep{zhang2025qronos}
&
Mismatched-input layer reconstruction,
$\|\mW_\ell\mX_\ell-
\widehat{\mW}_\ell
\widetilde{\mX}_\ell\|_F^2$.
&
Taking
$\widehat{\mW}_\ell=
\widehat{\mW}_\ell(\mG_\ell)$
gives $f_{\rm M}$ above.
Qronos uses causal diffusion to future
coordinates, but its full procedure also
corrects for the upstream-input mismatch.
\\
\bottomrule
\end{tabular}
\end{table*}

The table compares objectives and compensation
mechanisms, not identical feasible sets or
algorithms. In particular, Qronos's causal
diffusion should not be mistaken for the
absence of triangular feedback. Its matrix
denoted $G$ is a cross-Gram matrix between
upstream-quantized and full-precision inputs,
not our unit lower-triangular $\mG_\ell$.
If feedback matrices across multiple layers
are optimized jointly, then
$\widetilde{\mX}_\ell$ depends on earlier
feedback choices and is no longer fixed
layerwise context.

\section{Proofs and Matrix-Design Details}
\label{app:proofs}

Throughout this section, $\mH\in\R^{n\times n}$ is symmetric
positive definite, $\mR\in\R^{n\times n}$ is orthogonal, and
$\mG\in\R^{n\times n}$ is unit lower triangular. The results
concern the trace objective for a specified metric $\mH$.

\subsection{Optimal feedback for a fixed rotation}
\label{app:fixed-rotation}

Fix $\mR$ and set
\[
    \mA=\mR\mH\mR^\top\succ\bm 0.
\]
We first consider
\[
    \min_{\mG\ {\rm unit\ lower\ triangular}}
    \tr(\mG^\top\mA\mG).
\]
Because $\mA\succ\bm 0$, the objective is strictly convex in
the free entries of $\mG$. At its unique minimizer, stationarity
with respect to those entries requires $\mA\mG$ to be upper
triangular. Consequently, $\mG^\top\mA\mG$ is both upper
triangular and symmetric, and is therefore diagonal.

Write the reverse LDL factorization as
\[
    \mA=\mU\mD\mU^\top,
\]
where $\mU$ is unit upper triangular and $\mD$ is positive
diagonal. The matrix $\mG^\star=\mU^{-\top}$ is unit lower
triangular and satisfies
\[
    (\mG^\star)^\top\mA\mG^\star=\mD.
\]
It satisfies the stationarity condition and hence, by strict
convexity, is the unique minimizer. In particular,
\[
    \min_{\mG}\tr(\mG^\top\mA\mG)=\tr(\mD).
\]
Equivalently,
$\mA^{-1}=\mG^\star\mD^{-1}(\mG^\star)^\top$.

\subsection{Proof of Proposition~\ref{prop:joint_rg_global_optimum}}
\label{app:joint-proof}

For any feasible pair $(\mG,\mR)$, define
\[
    \mB=\mG^\top\mR\mH\mR^\top\mG.
\]
The matrix $\mB$ is positive definite. Since $\mG$ is unit
lower triangular and $\mR$ is orthogonal,
\[
    \det(\mB)
    =\det(\mG)^2\det(\mR)^2\det(\mH)
    =\det(\mH).
\]
The arithmetic geometric mean inequality applied to the
eigenvalues of $\mB$ therefore gives
\[
    \tr(\mB)
    \geq n\det(\mB)^{1/n}
    =n\det(\mH)^{1/n}.
\]
Let $\eta=\det(\mH)^{1/n}$. Take a geometric mean decomposition
\[
    \mH^{1/2}=\mQ\mT\mV^\top,
\]
where $\mQ,\mV$ are orthogonal and $\mT$ is upper triangular
with diagonal entries $\sqrt{\eta}$
~\citep{jiang2005gmd}. Set
\[
    \mR^\star=\mQ^\top,
    \qquad
    \mG^\star=\sqrt{\eta}\,\mT^{-\top}.
\]
Since $\mT^{-\top}$ is lower triangular with diagonal entries
$1/\sqrt{\eta}$, $\mG^\star$ is unit lower triangular. Also,
\[
    \mR^\star\mH(\mR^\star)^\top=\mT\mT^\top,
\]
and thus
\[
    (\mG^\star)^\top
    \mR^\star\mH(\mR^\star)^\top
    \mG^\star
    =\eta\mI.
\]
Its trace is $n\eta$, attaining the lower bound.

\subsection{Scope of the blockwise construction}
\label{app:blockwise-optimum}

The proposition allows an unrestricted orthogonal rotation.
In the experiments, the input coordinates are partitioned
into blocks $I_1,\ldots,I_K$, and rotation and feedback are
restricted to be block diagonal. Let $n_b=|I_b|$ and let
$\mH_{bb}=\mH[I_b,I_b]$. For this restricted problem, the
trace separates across blocks. If every $\mH_{bb}$ is
positive definite, the blockwise optimum is
\[
    T_{\rm block}^{\star}
    =
    \sum_{b=1}^{K}
    n_b\det(\mH_{bb})^{1/n_b}.
\]
This is the bound attained by independent blockwise
constructions. It is not, in general, the unrestricted
full-coordinate bound $n\det(\mH)^{1/n}$. If the design uses
damping, the reported trace and bound must both use the
same damped metric.

\section{Trace Optimality and Realized Quantization Error}
\label{app:trace-diagnostic}
 
We distinguish attainment of the block-restricted trace optimum from the
behavior of the corresponding finite-level quantizer. 
For layer $\ell$ and input-coordinate block $b$, let
\[
T_{\ell b}
=
\tr\!\left(
\mG_{\ell b}^{\top}\mR_{\ell b}
\mH_{\ell b}\mR_{\ell b}^{\top}\mG_{\ell b}
\right),
\;
B_{\ell b}
=
n_{\ell b}\det(\mH_{\ell b})^{1/n_{\ell b}},
\]
where $n_{\ell b}$ is the block width. Since the same block
design is applied to all $d_\ell$ output rows of layer $\ell$,
the whole-network trace ratio for each seed is
\[
\rho_{\rm trace}
=
\frac{\sum_{\ell}d_\ell\sum_{b\in\ell}T_{\ell b}}
     {\sum_{\ell}d_\ell\sum_{b\in\ell}B_{\ell b}}.
\]
 A value of one
verifies the restricted trace construction to numerical precision.

For each output row, let $\boldsymbol{q}_{b,r}$ denote the instantaneous
quantization-error vector generated within block $b$. Its shaped error in
original input coordinates is
$\boldsymbol{e}_{b,r}=\mR_b^\top\mG_b\boldsymbol{q}_{b,r}$.
The reported realized metric error aggregates
$\boldsymbol{e}_{b,r}^\top\mH_{bb}\boldsymbol{e}_{b,r}$ over blocks and rows.
This blockwise quadratic omits cross-block metric terms and is distinct from
the test logit MSE of the fully quantized network.
Table \ref{tab:trace-realized-diagnostic} reports these measurements under different set up.

\begin{table*}[t]
\centering
\caption{Trace optimality versus realized finite-level error on
MNIST at $\alpha=0$, averaged over 100 paired seeds. The trace
ratio is computed for each seed from row-weighted block traces
and determinant bounds, then averaged over seeds. Realized metric
error is the aggregate blockwise quadratic error; test logit MSE
is measured for the fully quantized network. Lower is better.}
\label{tab:trace-realized-diagnostic}
\resizebox{0.5\textwidth}{!}{
\footnotesize
\begin{tabular}{c l c r r}
\toprule
$M$ & Method & Trace ratio
    & Realized metric error & Test logit MSE\\
\midrule
2  & Covariance feedback
   & 1.98079 & \textbf{98.60880} & \textbf{1.69577}\\
   & Joint rotation-feedback
   & \textbf{1.00000} & 347.72978 & 9.21850\\
\midrule
4  & Covariance feedback
   & 1.98079 & \textbf{9.73280} & \textbf{0.08490}\\
   & Joint rotation-feedback
   & \textbf{1.00000} & 22.36973 & 0.15830\\
\midrule
8  & Covariance feedback
   & 1.98079 & \textbf{1.76087} & \textbf{0.01370}\\
   & Joint rotation-feedback
   & \textbf{1.00000} & 1.79542 & 0.01529\\
\midrule
16 & Covariance feedback
   & 1.98079 & 0.38108 & \textbf{0.00294}\\
   & Joint rotation-feedback
   & \textbf{1.00000} & \textbf{0.33103} & 0.00303\\
\bottomrule
\end{tabular}
}
\end{table*}
\begin{table*}[t]
\centering
\caption{Trace optimality versus realized finite-level error on
CIFAR-10 at $\alpha=0$, averaged over 100 paired seeds. Quantities
and aggregation follow Table~\ref{tab:trace-realized-diagnostic}.
Lower is better.}
\label{tab:cifar-trace-realized-diagnostic}
\resizebox{0.5\textwidth}{!}{
\small
\begin{tabular}{c l c r r}
\toprule
$M$ & Method & Trace ratio
    & Realized metric error & Test logit MSE\\
\midrule
2  & Covariance feedback
   & 1.13685 & \textbf{107.21706} & \textbf{1.83842}\\
   & Joint rotation-feedback
   & \textbf{1.00000} & 191.77801 & 3.52017\\
\midrule
4  & Covariance feedback
   & 1.13685 & \textbf{10.39471} & \textbf{0.10596}\\
   & Joint rotation-feedback
   & \textbf{1.00000} & 12.26181 & 0.12758\\
\midrule
8  & Covariance feedback
   & 1.13685 & \textbf{1.89431} & \textbf{0.01863}\\
   & Joint rotation-feedback
   & \textbf{1.00000} & 2.20850 & 0.02126\\
\midrule
16 & Covariance feedback
   & 1.13685 & \textbf{0.41161} & \textbf{0.00403}\\
   & Joint rotation-feedback
   & \textbf{1.00000} & 0.47938 & 0.00453\\
\bottomrule
\end{tabular}
}
\end{table*}

Across 100 paired seeds, joint rotation-feedback attains the
blockwise determinant bound to numerical precision on both
datasets. This trace ordering does not generally carry over to
finite-level performance. Covariance feedback has lower mean
test logit MSE at every tested resolution on both MNIST and
CIFAR-10. On MNIST at $M=16$, joint design has lower mean
realized blockwise metric error, but its mean test logit MSE
remains slightly higher. Thus, attainment of the trace bound,
realized blockwise metric error, and end-to-end logit MSE
remain distinct comparisons.


\section{Post-Training Experimental Protocols}
\label{app:ptq-protocols}

This appendix specifies the design and evaluation conventions for
the post-training quantization experiments. Throughout, $M$ denotes
the number of quantization levels. For a weight row,
$q_i=\widehat w_i-b_i$ is the instantaneous quantization error,
whereas $\widehat{\vw}-\vw=\mG\vq$ is the shaped weight error.
Quantization grids are fixed before causal feedback is applied.

\subsection{Synthetic networks}
\label{app:synthetic-protocol}

The synthetic experiment uses a five-linear-layer network with
widths $12\to16\to16\to12\to8\to3$. The first four layers have
$\tanh$ activations, and the output layer is linear. For each of
$100$ independently generated networks and datasets, we use $256$
calibration samples and an independent set of $1024$ evaluation
samples. The weights are drawn from zero-mean Gaussian
distributions with standard deviations proportional to the
inverse square root of the layer input width. Inputs are generated
from a four-dimensional latent Gaussian variable through a
mixing matrix fixed within each trial, with additive Gaussian
noise.

For output row $r$ of layer $\ell$, let $\bar{\mC}_\ell$ and
$\mH_{{\rm Jac},\ell,r}^{\rm row}$ be the sample-averaged input
covariance and row-wise Jacobian metric defined in
Section~\ref{sec:noise-shaping}. The same scale is applied to
all rows of a layer:
\[
    s_\ell
    =
    \frac{\tr(\bar{\mC}_\ell)}
    {\max\!\left\{
       d_\ell^{-1}\sum_{r=1}^{d_\ell}
       \tr(\mH_{{\rm Jac},\ell,r}^{\rm row}),
       10^{-30}
    \right\}}.
\]
For each row, feedback is designed from
\[
\begin{aligned}
    \mA_{\ell,r}(\alpha)
    &=
    \bar{\mC}_\ell
    +\alpha s_\ell\mH_{{\rm Jac},\ell,r}^{\rm row},\\
    \widetilde{\mA}_{\ell,r}(\alpha)
    &=
    \mA_{\ell,r}(\alpha)
    +10^{-5}
      \frac{\tr(\mA_{\ell,r}(\alpha))}
           {d_{\ell-1}}\mI .
\end{aligned}
\]
The second line is the trace-scaled ridge used before computing
the feedback matrix.

For each weight row and $M$, the grid consists of $M$ evenly
spaced levels between the minimum and maximum entries of the
original row. It is fixed across
$\alpha\in\{0,0.01,0.03,0.1,0.2,0.4,0.6,0.8,0.99\}$.
Quantizer inputs outside the grid range map to an endpoint.
The same $256$ calibration samples are used to estimate the
design metrics and to select $\alpha$ by output MSE relative
to the full-precision network. The independent evaluation set
is used only to report test output MSE. Thus, unlike the
trained-network experiments below, this synthetic experiment
does not use a disjoint $\alpha$-selection set.

\subsection{Trained MNIST and CIFAR-10 networks}
\label{app:mlp-protocol}

The MNIST network has widths
$784\to128\to128\to64\to32\to10$; the CIFAR-10 network has
widths $3072\to128\to128\to64\to32\to10$. Inputs are
flattened, the first four linear layers have $\tanh$
activations, and the output layer is linear. The reported
results are averaged over $100$ independently trained
checkpoints. In training,
$2048$ training samples are reserved for metric calibration 
and $512$  training samples are reserved separately for $\alpha$ selection;
the remaining training set are used to fit the checkpoint.

Input coordinates within each weight row are partitioned
into consecutive blocks of at most $32$. A separate feedback
matrix is designed for each block. The RHT and joint methods
also use a separate rotation for each block. Each block's
design is shared across the output rows of its layer.

For block $I_b$, let $\mX_b$ contain the corresponding
calibration activations and define
\[
\begin{aligned}
\mC_b
    &=
    \frac{1}{N_{\rm cal}}\mX_b\mX_b^\top,\\
    \mH_{{\rm Jac},b}^{\rm iso}
    &=
    \frac{1}{N_{\rm cal}}
    \sum_{t=1}^{N_{\rm cal}}
    \frac{\|\mJ_{\ell,t}\|_F^2}{d_\ell}\,
    \vx_{\ell,t,b}\vx_{\ell,t,b}^{\top}.
\end{aligned}
\]

The effective design metric is constructed separately for
each block:
\[
\begin{aligned}
    \mA_b(\alpha)
    &=
    \mC_b+
    \alpha\,
    \frac{\tr(\mC_b)}
         {\max\{\tr(\mH_{{\rm Jac},b}^{\rm iso}),10^{-30}\}}
    \mH_{{\rm Jac},b}^{\rm iso},\\
    \widetilde{\mH}_b(\alpha)
    &=
    \mA_b(\alpha)
    +\lambda
      \frac{\tr(\mA_b(\alpha))}{|I_b|}\mI .
\end{aligned}
\]
We set $\lambda=10^{-3}$. 
The covariance-based methods use $\alpha=0$; 
the Jacobian-aware methods select $\alpha$ among the candidates in $\{0,0.03,0.1,0.3,1\}$.
The joint method applies
its rotation-feedback construction to the damped block
metric $\widetilde{\mH}_b(\alpha)$; its determinant bound
therefore refers to that metric.

Full-precision network Jacobians on the metric-calibration
set are used to construct the candidates. Jacobians on the
disjoint selection set are used to score them. Specifically,
the selection criterion is the full first-order
network-output quadratic
\[
    D_{\rm sel}^{(1)}
    =
    \frac{1}{N_{\rm sel}d_L}
    \sum_{t=1}^{N_{\rm sel}}
    \left\|
       \sum_{\ell=1}^{L}
       \mJ_{\ell,t}
       (\widehat{\mW}_\ell-\mW_\ell)
       \vx_{\ell,t}
    \right\|_2^2.
\]
This evaluates the resulting candidate weights; it is not
the row-isotropic trace objective used to construct them.
Final test logit MSE is measured from the fully quantized
network. Accuracy drop is calculated relative to each
checkpoint's full-precision accuracy, in percentage points,
before averaging over checkpoints.

Within each block, each rotated weight row receives an
evenly spaced $M$-level grid spanning that row's minimum
and maximum before feedback. Inputs outside the grid range
map to an endpoint. For an RHT block whose size is not a
power of two, the implementation uses a random signed
identity instead of a Hadamard transform.

\subsection{GPT-2 projection weights}
\label{app:gpt2-protocol}

The reported experiment uses full-precision embedding mode.
We quantize four attention and MLP projection matrices in
each of the $12$ GPT-2 transformer blocks:
\texttt{attn.c\_attn}, \texttt{attn.c\_proj},
\texttt{mlp.c\_fc}, and \texttt{mlp.c\_proj}.
Thus, $48$ projection matrices are quantized. Token and
positional embeddings remain in full precision, as does the
output head tied to the token embedding. Other parameters
outside these projection matrices are not quantized. The
projection weights are processed in input-coordinate blocks
of size $32$.

Metric calibration, $\alpha$ selection, and test evaluation
use non-overlapping length-$64$ token windows from the
WikiText-2 raw train, validation, and test splits,
respectively. The script tokenizes nonempty text rows,
inserts a separator between rows, and takes consecutive
windows from the resulting token stream. Each window
contributes $63$ scored next-token positions.

For every calibration batch, the script draws one
Rademacher vector over the vocabulary at each scored
position, with entries scaled by the inverse square root
of the vocabulary size. Backpropagating the resulting
logit probe through the full-precision model gives a
downstream gradient at each selected projection output.
The squared gradient is averaged over that projection's
output coordinates to obtain one sensitivity value per
scored token. The same probe supplies gradients for the
selected projections in that calibration batch.

For each projection and input-coordinate block, the
script accumulates a sample-averaged activation covariance
$\mC_b$ and a sensitivity-weighted activation covariance
$\mH_b$. It then sets
\[
    s_b
    =
    \frac{\tr(\mC_b)}
         {\max\{\tr(\mH_b),10^{-30}\}},
    \qquad
    \mA_b(\alpha)
    =
    \mC_b+\alpha s_b\mH_b,
\]
and uses the damped metric
\[
    \widetilde{\mH}_b(\alpha)
    =
    \mA_b(\alpha)
    +\lambda
      \frac{\tr(\mA_b(\alpha))}{|I_b|}\mI .
\]
We set $\lambda=10^{-3}$.
This is not the full downstream-Jacobian quadratic but a stochastic row-isotropic design surrogate. 
Feedback and, where applicable, rotation are designed independently
within each block.

Each rotated weight row receives a fixed, evenly spaced
$M$-level grid between its pre-feedback minimum and maximum.
Quantizer inputs outside that range saturate at an endpoint.
The Jacobian-aware methods select $\alpha$ by validation
logit MSE against the full-precision model after quantizing
all selected projection matrices. The covariance-feedback
baseline uses fixed-order feedback rather than the full
published GPTQ implementation.

Test perplexity is computed from next-token cross-entropy,
and test logit MSE is measured relative to the
full-precision model over the same scored positions.
The reported evaluation uses $128$ length-$64$ test
windows, comprising $8064$ scored next-token positions,
in a single run. These values describe the evaluated
windows rather than the entire WikiText-2 test corpus.
We use $32$ length-$64$ windows from the training split for metric calibration,
$32$ from the validation split for $\alpha$ selection, 
and $128$ from the test split for evaluation. 
The reported run uses seed $0$, input-coordinate blocks of size $32$, 
and $M\in\{2,4,8,16\}$ quantization levels. 
For the Jacobian-aware methods, 
candidate values are $\alpha\in\{0,0.03,0.1,0.2,0.8,0.99\}$.


\section{Architecture-dependent treatment of downstream curvature.}
The six baselines combine two distortion criteria, Local and Logit,
with three coordinate-basis and feedback choices, GPTQ, RHT, and
Joint. The interpretation of the Logit criterion differs between
the LLM and CIFAR-10 CNN experiments.

Let $a_{\ell,s}$ denote the vectorized preactivation of layer $\ell$
for calibration example $s$, and let $f_s$ collect the final logits.
Define
\[
J_{\ell,s}
=
\frac{\partial f_s}{\partial a_{\ell,s}^{\top}},
\qquad
C_{\ell,s}=J_{\ell,s}^{\top}J_{\ell,s}.
\]
For a local perturbation $\delta a_{\ell,s}$, the first-order
approximation to the squared logit perturbation is
\[
\|\delta f_s\|_2^2
\approx
\delta a_{\ell,s}^{\top}
C_{\ell,s}
\delta a_{\ell,s}.
\]

In the LLM experiments, the downstream curvature is approximated
by a diagonal matrix,
\[
C_{\ell,s}
\approx
\operatorname{Diag}(\widehat d_{\ell,s}).
\]
The diagonal sensitivities are obtained by deterministic
operation-wise propagation. For an operation $y=\phi(x)$ with
Jacobian $B=\partial y/\partial x^\top$, the propagation rule is
\[
\widehat d_x
=
\operatorname{diag}
\left(
B^\top \operatorname{Diag}(\widehat d_y) B
\right)
=
(B\odot B)^\top\widehat d_y.
\]
This procedure discards intermediate off-diagonal curvature and
cross-path covariance terms. Consequently, $\widehat d_{\ell,s}$
is generally an approximation to
$\operatorname{diag}(J_{\ell,s}^{\top}J_{\ell,s})$, rather than its
exact evaluation.

The final RMSNorm and language-model head are treated analytically
using the full vocabulary. This gives an exact initialization of
the head sensitivity in the differentiable real-arithmetic model.
However, exact head initialization does not make the preceding
diagonal propagation exact.

In contrast, the CIFAR-10 CNN experiments retain the full
$J_{\ell,s}^{\top}J_{\ell,s}$ without diagonalization.
Writing
\[
\delta a_{\ell,s}
=
B_{\ell,s}\delta\theta_\ell,
\]
where $\delta\theta_\ell$ stacks the perturbed parameters of
layer $\ell$, the exact layerwise weight-space curvature is
\[
\mathcal H_\ell^{\mathrm{exact}}
=
\frac{1}{NC}
\sum_{s=1}^{N}
B_{\ell,s}^{\top}
J_{\ell,s}^{\top}J_{\ell,s}
B_{\ell,s},
\qquad C=10.
\]
Equivalently, with the full layer-weight Jacobian
\[
\mathcal J_{\ell,s}
=
\frac{\partial f_s}{\partial\theta_\ell^\top}
=
J_{\ell,s}B_{\ell,s},
\]
we have
\[
\mathcal H_\ell^{\mathrm{exact}}
=
\frac{1}{NC}
\sum_{s=1}^{N}
\mathcal J_{\ell,s}^{\top}\mathcal J_{\ell,s}.
\]
The exact Jacobian factor is retained, including within-layer
cross-filter and spatial-path interactions; no diagonal,
Monte Carlo, or rank-truncation approximation is introduced
in this curvature construction.

Here, ``exact'' refers to the Jacobian-based quadratic evaluated
at the reference network. It does not imply an exact description
of the nonlinear network loss after finite quantization, nor
an unrestricted joint optimization over all network parameters.

\section{Feedback Optimization in the Toy QAT Experiment}
\label{app:qat-details}

\subsection{Data, model, and quantization grids}
The student is a two-layer regression network with dimensions
$4\to6\to3$ and a $\tanh$ hidden activation. Each of 100 paired trials
generates an independent teacher network, correlated four-dimensional
inputs, and targets with independent additive Gaussian noise of standard
deviation $0.03$. Training, validation, and test sets contain 160, 80,
and 256 input-target pairs. Within a trial, all methods share the data,
full-precision student checkpoint, and quantization grids. The student
is first fitted for 160 full-batch Adam steps at learning rate $0.025$.

Both weight matrices are quantized at $M\in\{2,4,8\}$ levels. Each row uses an evenly spaced grid between the
minimum and maximum of that row in the fitted full-precision checkpoint.
The grids are held fixed during subsequent training, even when shadow
weights change. Nearest-grid rounding saturates at the grid endpoints.

\subsection{Causal quantizer and feedback initialization}
A unit lower-triangular $\mG_\ell$ is shared across the output rows of
layer $\ell$, with dimensions $4\times4$ and $6\times6$ for the two
layers. For row $r$, the hard-quantized forward pass follows the
input-coordinate order
\[
\begin{aligned}
 b_{\ell,r,i}&=w_{\ell,r,i}+\sum_{j<i}g_{\ell,ij}q_{\ell,r,j},\\
 \widehat w_{\ell,r,i}&=\mathcal Q_{M,\ell,r}(b_{\ell,r,i}),\\
 q_{\ell,r,i}&=\widehat w_{\ell,r,i}-b_{\ell,r,i}.
\end{aligned}
\]
Although feedback coefficients are shared across rows, each row has
its own instantaneous-error sequence.

The covariance-feedback initialization is computed from the empirical
input second moment at the full-precision checkpoint. For the second
layer, the inputs are the full-precision hidden activations. The metric
used for this initialization receives the ridge
\[
 \mH_\ell\longleftarrow
 \mH_\ell+0.02\,\frac{\operatorname{tr}(\mH_\ell)}{n_\ell}\mI.
\]
This ridge affects covariance-based initialization only. The
unregularized QAT results below set the feedback-training penalty to
zero; feedback entries are still clipped after updates.

\subsection{Compared methods and surrogate gradients}
RTN uses identity feedback without training. Covariance-feedback PTQ
uses the covariance-designed feedback with the full-precision network
fixed. Feedback-only tuning starts from covariance feedback and updates
only $\mG$, leaving weights fixed. Weight-only QAT starts
from $\mG_\ell=\mI$, holds $\mG$ fixed, and updates shadow weights. Joint QAT also starts from $\mG_\ell=\mI$ and the
same full-precision checkpoint, but updates the strictly
lower-triangular feedback entries as well as weights.
Thus, the two QAT arms in the selected main-text comparison have a
common identity-feedback initialization. The full comparison in
Table~\ref{tab:qat-all-arms} also reports covariance-initialized QAT
arms for context.

All tuning arms use 100 full-batch Adam steps. The learning rate is
$0.015$ for trainable weights and $0.009$ for trainable
feedback entries. The forward pass uses hard-quantized weights. In the
backward pass, the scalar quantizer has the identity straight-through
derivative, including at saturated inputs. Previously generated
$q_{\ell,r,j}$ values are treated as constants when differentiating
later quantizer inputs. Writing
$S_{\ell,r,i}=\partial\mathcal L/\partial\widehat w_{\ell,r,i}$
for the gradient of the hard-quantized network's mean-squared error,
the implemented surrogate gradients are
\[
 \widetilde{\frac{\partial\mathcal L}{\partial w_{\ell,r,i}}}
 =S_{\ell,r,i},\qquad
 \widetilde{\frac{\partial\mathcal L}{\partial g_{\ell,ij}}}
 =\sum_r S_{\ell,r,i}q_{\ell,r,j},\quad j<i.
\]
Here the loss is the mean of squared errors over samples and output
coordinates. After each feedback update, the strictly lower-triangular
entries are clipped to $[-2,2]$, and the diagonal and upper triangle
are reset to those of the identity. There is no $\mG$-training
regularizer in these results.

Every ten steps, the hard-quantized network is evaluated on the
validation set. The checkpoint with lowest validation MSE is retained,
and its test MSE is computed on the independent test set. Test targets
are not used to select checkpoints.

\subsection{Complete comparison}
\begin{table*}[t]
\centering
\caption{Mean test MSE over 100 paired synthetic regression trials
without a feedback-training regularizer. Lower is better. Boldface
indicates the lowest mean among quantized methods in each column.
Feedback-only tuning starts from covariance feedback; the
identity-initialized QAT arms start from $\mG=\mI$.}
\label{tab:qat-all-arms}
\small
\begin{tabular}{lccc}
\toprule
Method & $M=2$ & $M=4$ & $M=8$\\
\midrule
Full precision & 0.01391 & 0.01391 & 0.01391\\
RTN & 0.63739 & 0.06576 & 0.02395\\
Covariance-feedback PTQ & 0.48603 & 0.05473 & 0.02078\\
Feedback-only tuning; network fixed & 0.28594 & 0.04055 & 0.01899\\
Weight-only QAT; $\mG=\mI$ fixed & 0.25458 & 0.04118 & 0.01914\\
Joint QAT; $\mG_0=\mI$ & 0.24524 & 0.04013 & 0.01784\\
Weight-only QAT; covariance $\mG$ fixed
  & 0.23674 & \textbf{0.03800} & \textbf{0.01763}\\
Joint QAT; covariance $\mG_0$
  & \textbf{0.23228} & 0.03844 & 0.01766\\
\bottomrule
\end{tabular}
\end{table*}

Feedback-only tuning lowers mean test MSE relative to
covariance-feedback PTQ at each tested resolution while keeping the
fitted network fixed. Among the identity-initialized QAT arms, joint
training has lower mean test MSE than weight-only QAT at each tested
resolution. The additional covariance-initialized arms show that
initialization also matters: joint training does not uniformly
outperform QAT with covariance feedback held fixed.

\section{Additional Experimental Analyses}
\label{sec:additional_results}

These experiments complement the main-text comparisons under different architectures, grids, constraints, and evaluation protocols. EFQ denotes causal error-feedback quantization. Local EFQ uses a local input-coordinate criterion, whereas Obj.-EFQ uses the downstream criterion named in each experiment. EFQ+RHT applies randomized orthogonal preprocessing before feedback; joint $R,G$ denotes the corresponding joint-design procedure. Metric names in parentheses specify the criterion used; downstream formulations differ across experiments. Accuracy is reported in percent throughout, and reported standard deviations are sample standard deviations.

\subsection{MNIST: Quantization-Procedure Comparison}
\label{sec:exp_mnist}

\paragraph{Experimental setup.}
We quantize three pretrained ReLU MLPs with architecture $784\to64\to32\to10$ and seeds $\{7,8,42\}$ at $M\in\{2,4,8,16\}$ levels, without retraining. For each seed and bit width, damping is selected on a disjoint 5,000-example training holdout; the selected quantizer is then rebuilt from the original model using the training data and evaluated on the 10,000-example test set. Local EFQ uses asymmetric affine grids and full-precision biases, whereas unconstrained Obj.-EFQ (Logit-MSE) uses symmetric Sigma--Delta grids and joint weight--bias quantization. Table~\ref{tab:mnist_main} therefore compares complete quantization procedures rather than an isolated change of metric or constraint.

\begin{table*}[t]
\centering
\caption{MNIST test accuracy (\%, mean $\pm$ sample standard deviation over three pretrained seeds). Bold marks the higher mean at each bit width. The two procedures differ in quantization grid and bias treatment.}
\label{tab:mnist_main}
\small
\setlength{\tabcolsep}{4pt}
\renewcommand{\arraystretch}{1.10}
\begin{adjustbox}{max width=\textwidth}
\begin{tabular}{@{}lcccc@{}}
\toprule
Method & 1 bit ($M=2$) & 2 bits ($M=4$)
& 3 bits ($M=8$) & 4 bits ($M=16$) \\
\midrule
Local EFQ
& $10.750\pm1.041$
& $95.063\pm0.285$
& $\mathbf{97.167\pm0.299}$
& $97.237\pm0.265$ \\
Obj.-EFQ (Logit-MSE), unconstrained
& $\mathbf{85.560\pm3.147}$
& $\mathbf{95.607\pm0.948}$
& $97.160\pm0.210$
& $\mathbf{97.257\pm0.337}$ \\
\bottomrule
\end{tabular}
\end{adjustbox}
\end{table*}

\paragraph{Results and analysis.}
Obj.-EFQ has higher mean accuracy at one and two bits, while the two procedures are close at three and four bits relative to the reported across-seed variation. The one-bit means differ substantially ($85.560\%$ versus $10.750\%$), but the simultaneous changes in grid construction and bias treatment prevent attributing this difference solely to the downstream criterion or removal of the no-overload constraint.

\subsection{CIFAR-10: Layerwise Curvature and Rotation}
\label{sec:exp_cifar10}

\paragraph{Experimental setup.}
We evaluate layerwise PTQ on CIFAR-10 using its 50,000 training and 10,000 test images. For each seed in $\{42,33,8\}$, a stratified 5,000-image validation set is held out, and 512 calibration images are sampled without replacement from the remaining 45,000 training images. Inputs use per-channel normalization; teacher training additionally uses random crops with four-pixel padding and horizontal flips. The CNN has seven convolutions, each followed by batch normalization and ReLU, with $2\times2$ max pooling after convolutions 3, 5, and 7, global average pooling, and a $256\to10$ classifier. Folding batch normalization gives 292,826 affine parameters. Teachers are trained for 200 epochs with SGD, Nesterov momentum $0.9$, weight decay $5\times10^{-4}$, batch size 256, learning rate $0.1$, five-epoch linear warm-up, cosine decay, and BF16 mixed precision; checkpoint selection uses validation accuracy, then NLL. The FP32 teachers attain $89.77\pm0.14\%$ test accuracy. All fused weights and biases are quantized without fine-tuning using per-output-row symmetric grids at $M\in\{2,4,8,16\}$; each clipping radius is selected from $\{1.00,0.95,\ldots,0.50\}$ times the row maximum by minimizing rounding MSE. We cross Local affine-output reconstruction, Obj.-diag exact downstream derivatives with cross-output curvature discarded, and Obj.-exact complete within-layer curvature $H=J^\top J/(NC)$ with EFQ, EFQ+RHT, and joint $R,G$. Here $N$ is the calibration size and $C=10$; Obj.-exact retains cross-filter and spatial couplings but excludes cross-layer curvature. Joint design equalizes a loaded column-contracted metric before feedback. The GPTQ-style recurrence uses 128-coordinate computational blocks and rebuilds dual factors every 128 coordinates; two numerical retries with tenfold loading increments are allowed, but none is used. All 108 configurations are evaluated on the full training and test sets using dense FP32 reconstruction, recording accuracy, NLL, teacher-to-student KL, and per-class logit MSE. Table~\ref{tab:uniform-cifar10-ptq} reports three-seed test means and standard deviations.

\begin{table*}[t]
\centering
\caption{Layerwise PTQ on CIFAR-10. Test accuracy (\%) and logit MSE are reported as mean $\pm$ sample standard deviation over seeds $\{42,33,8\}$. All fused weights and biases are quantized. Bold marks the highest mean accuracy at each bit width; the FP32 reference attains $89.77\pm0.14\%$.}
\label{tab:uniform-cifar10-ptq}
\small
\setlength{\tabcolsep}{4pt}
\renewcommand{\arraystretch}{1.10}
\begin{adjustbox}{max width=\textwidth}
\begin{tabular}{@{}lcccccccc@{}}
\toprule
& \multicolumn{2}{c}{1 bit ($M=2$)}
& \multicolumn{2}{c}{2 bits ($M=4$)}
& \multicolumn{2}{c}{3 bits ($M=8$)}
& \multicolumn{2}{c}{4 bits ($M=16$)} \\
\cmidrule(lr){2-3}\cmidrule(lr){4-5}
\cmidrule(lr){6-7}\cmidrule(lr){8-9}
Method
& Acc.\ $\uparrow$ & Logit MSE $\downarrow$
& Acc.\ $\uparrow$ & Logit MSE $\downarrow$
& Acc.\ $\uparrow$ & Logit MSE $\downarrow$
& Acc.\ $\uparrow$ & Logit MSE $\downarrow$ \\
\midrule
Local EFQ
& $9.68\pm0.62$ & $6556\pm1780$
& $17.96\pm3.72$ & $37.70\pm2.92$
& $49.10\pm5.29$ & $27.38\pm4.22$
& $73.70\pm5.33$ & $11.58\pm2.46$ \\
Local EFQ+RHT
& $9.99\pm0.14$ & $6672\pm2899$
& $17.63\pm2.63$ & $38.83\pm2.08$
& $52.36\pm6.38$ & $26.02\pm5.12$
& $74.59\pm5.12$ & $10.91\pm3.96$ \\
Local joint $R,G$
& $9.53\pm0.89$ & $7238\pm6271$
& $17.94\pm1.48$ & $39.03\pm1.71$
& $48.50\pm9.55$ & $28.12\pm9.82$
& $75.18\pm5.28$ & $10.60\pm3.73$ \\
\midrule
Obj.-EFQ (diag)
& $10.31\pm0.14$ & $2103\pm1416$
& $\mathbf{28.11}\pm10.94$ & $40.52\pm10.52$
& $54.51\pm10.95$ & $23.51\pm5.20$
& $70.96\pm12.53$ & $13.54\pm8.25$ \\
Obj.-EFQ+RHT (diag)
& $\mathbf{11.79}\pm1.04$ & $1972\pm660$
& $18.71\pm4.48$ & $42.18\pm3.88$
& $45.86\pm4.16$ & $27.92\pm2.58$
& $76.16\pm5.20$ & $9.86\pm3.64$ \\
Obj. joint $R,G$ (diag)
& $10.53\pm0.99$ & $3508\pm1481$
& $17.55\pm2.19$ & $39.44\pm5.07$
& $46.16\pm17.76$ & $29.17\pm10.04$
& $74.83\pm4.86$ & $9.75\pm2.80$ \\
\midrule
Obj.-EFQ (exact)
& $10.94\pm2.29$ & $2149\pm2077$
& $15.17\pm5.32$ & $44.82\pm9.77$
& $54.55\pm11.28$ & $19.57\pm3.92$
& $\mathbf{82.40}\pm0.55$ & $5.84\pm0.43$ \\
Obj.-EFQ+RHT (exact)
& $8.90\pm0.31$ & $3163\pm975$
& $23.12\pm6.16$ & $37.95\pm4.35$
& $\mathbf{55.73}\pm10.52$ & $21.67\pm1.90$
& $81.90\pm0.17$ & $6.09\pm0.62$ \\
Obj. joint $R,G$ (exact)
& $10.23\pm1.49$ & $2558\pm2271$
& $15.92\pm4.18$ & $40.53\pm0.47$
& $55.28\pm4.28$ & $19.56\pm2.06$
& $81.00\pm1.44$ & $6.31\pm0.42$ \\
\bottomrule
\end{tabular}
\end{adjustbox}
\end{table*}

\paragraph{Results and analysis.}
The clearest benefit of complete within-layer curvature appears at four bits: Obj.-EFQ (exact) attains $82.40\%$ accuracy and logit MSE $5.84$, exceeding Local EFQ by $8.70$ percentage points and Obj.-EFQ (diag) by $11.44$ points, although a $7.37$-point gap to the teacher remains. At three bits, all exact variants attain approximately $55\%$ accuracy; the RHT variant has the highest accuracy ($55.73\%$), whereas joint $R,G$ has the lowest logit MSE ($19.56$). Thus, output fidelity and classification accuracy need not rank methods identically. Rotation gains depend on the metric and precision: RHT helps Local EFQ at three and four bits but hurts Obj.-EFQ (diag) at two and three bits, and unrotated exact EFQ outperforms both rotated exact variants at four bits. Joint design is not uniformly best, consistent with its optimization of a column-contracted metric rather than the complete coupled objective. At one and two bits, even the best accuracies are only $11.79\%$ and $28.11\%$; the mixed rankings are consistent with limited predictive power of calibration-point quadratic models under large perturbations, rather than a general advantage of local curvature.

\subsection{MODNet: Regression Comparison}
\label{sec:exp_modnet}

\paragraph{Experimental setup.}
We evaluate three ReLU MLPs with architecture $1000\to128\to64\to32\to1$ on the MODNet refractive-index task, using a fixed train/model-validation/quantization-validation/test split of $2614/402/402/604$ samples. Affine weights and biases are quantized without fine-tuning at the values of $M$ shown in Table~\ref{tab:modnet_main}. Damping is selected separately for each seed, method, and resolution using quantization-validation MSE; the test set is reserved for final evaluation. The table compares Local EFQ (GPTQ-style), Obj.-EFQ (multilayer), and full precision. The multilayer criterion here is not assumed to be identical to the row-isotropic Jacobian metric in the main-text MLP experiment.

\begin{table*}[t]
\centering
\caption{MODNet test MSE (mean $\pm$ sample standard deviation over three seeds; lower is better). Bold marks the lower mean among the two quantized procedures at each value of $M$.}
\label{tab:modnet_main}
\small
\setlength{\tabcolsep}{4pt}
\renewcommand{\arraystretch}{1.10}
\begin{adjustbox}{max width=\textwidth}
\begin{tabular}{@{}lcccc@{}}
\toprule
Method & $M=2$ & $M=4$ & $M=8$ & $M=16$ \\
\midrule
Local EFQ (GPTQ-style)
& $\mathbf{0.287104\pm0.240392}$
& $\mathbf{0.015371\pm0.002331}$
& $0.014062\pm0.000881$
& $\mathbf{0.013476\pm0.001157}$ \\
Obj.-EFQ (multilayer)
& $0.291390\pm0.226718$
& $0.016781\pm0.004631$
& $\mathbf{0.013747\pm0.001241}$
& $0.013535\pm0.001172$ \\
\midrule
Full precision
& \multicolumn{4}{c}{$0.013615\pm0.001164$} \\
\bottomrule
\end{tabular}
\end{adjustbox}
\end{table*}

\paragraph{Results and analysis.}
Local EFQ has lower mean test MSE at $M=2,4,16$, whereas Obj.-EFQ has lower mean MSE at $M=8$. At $M=16$, both quantized means are close to the full-precision mean relative to the three-seed variation; their slightly smaller means do not establish an improvement over full precision. These conclusions concern only the displayed procedures, not the QEP variants from the broader experimental protocol.

\subsection{Qwen3-8B: Projection-Level Diagnostic}
\label{sec:exp_qwen8_projection}

\paragraph{Experimental setup.}
This sequential-replay experiment compares Local EFQ, Local EFQ+RHT, and Local joint $R,G$ on all 252 Qwen3-8B attention and MLP projections using whole-row asymmetric affine grids at one to four bits. Embeddings, normalization layers, and the language-model head remain in BF16. Calibration uses 128 WikiText-2 sequences of length 2048, with 32 additional validation sequences; evaluation uses 146 complete test sequences. For each projection, the original and quantized weights are applied to the same live replay input $\mathbf X$, giving
\[
\operatorname{MSE}_{\mathrm{out}}
=
\frac{\|\mathbf X\mathbf W^\top
-\mathbf X\widehat{\mathbf W}^{\top}\|_F^2}
{N d_{\mathrm{out}}},
\]
where $N$ is the number of probe inputs. Replay inputs reflect preceding quantization decisions and can therefore differ across methods. Table~\ref{tab:qwen8_projection} reports projection-level MSE, not final-logit MSE. Its sequence length, grid, and evaluation protocol differ from the main-text Qwen3-8B experiment, so the two BF16 perplexities are not directly comparable.

\begin{table*}[t]
\centering
\caption{Qwen3-8B sequential-replay diagnostic on WikiText-2 with length-2048 sequences. MSE statistics summarize direct projection-output error across 252 projections, not final-logit error. ``Wins'' counts the projections with the lowest MSE at each bit width. Bold marks the best quantized value in each column at that bit width.}
\label{tab:qwen8_projection}
\small
\setlength{\tabcolsep}{4pt}
\renewcommand{\arraystretch}{1.10}
\begin{adjustbox}{max width=\textwidth}
\begin{tabular}{@{}clrrrrr@{}}
\toprule
Bits & Method & PPL $\downarrow$
& Mean MSE $\downarrow$ & Median MSE $\downarrow$
& P95 MSE $\downarrow$ \\
\midrule
1 & Local EFQ
  & $1.0993{\times}10^{8}$ & 39.3955
  & 0.16201 & 42.4530 \\
1 & Local EFQ+RHT
  & $\mathbf{3.8108{\times}10^{5}}$ & 5.4685
  & \textbf{0.13079} & \textbf{9.9605} \\
1 & Local joint $R,G$
  & $1.0715{\times}10^{8}$ & \textbf{2.7440}
  & 0.39303 & 15.2230 \\
\midrule
2 & Local EFQ
  & 26488.66 & 1.9854 & 0.16360
  & 11.3747 \\
2 & Local EFQ+RHT
  & \textbf{833.66} & 0.85167
  & 0.10521 & 4.8990 \\
2 & Local joint $R,G$
  & 2584.97 & \textbf{0.80394}
  & \textbf{0.10230} & \textbf{4.8091} \\
\midrule
3 & Local EFQ
  & 14.1148 & 0.14179
  & 0.01630 & 0.83881 \\
3 & Local EFQ+RHT
  & \textbf{11.6523} & \textbf{0.05658}
  & \textbf{0.01005} & \textbf{0.33736} \\
3 & Local joint $R,G$
  & 11.9912 & 0.06643
  & 0.01185 & 0.42556 \\
\midrule
4 & Local EFQ
  & 10.3875 & 0.02881
  & 0.003204 & 0.17395 \\
4 & Local EFQ+RHT
  & 10.0141 & \textbf{0.01185}
  & \textbf{0.002070} & \textbf{0.06981} \\
4 & Local joint $R,G$
  & \textbf{10.0004} & 0.01372
  & 0.002380 & 0.08593 \\
\midrule
-- & BF16 & 9.7245 & 0 & 0 & 0 \\
\bottomrule
\end{tabular}
\end{adjustbox}
\end{table*}

\paragraph{Results and analysis.}
Local EFQ+RHT has lower mean projection-output MSE than joint $R,G$ at three and four bits and wins more projections. At two bits, joint $R,G$ has lower mean MSE and more wins, yet its PPL is higher ($2584.97$ versus $833.66$). At four bits, joint $R,G$ instead has slightly lower PPL ($10.0004$ versus $10.0141$) despite higher projection-output MSE. These diagnostics need not rank quantizers alike, especially when replay inputs are method-dependent. The one-bit results also require caution because the asymmetric two-level grid is not a symmetric sign quantizer.

\subsection{Qwen3-8B: Additional Metrics and Backends}
\label{sec:exp_qwen8_factorial}

\paragraph{Experimental setup.}
We extend the main-text Qwen3-8B experiment with Guided-shared and Model-Fisher-diag metrics and coordinate descent (CD), giving the 15 procedures in Table~\ref{tab:qwen8_factorial}. Calibration uses 128 WikiText-2 sequences of length 512, and evaluation covers 298,424 scored test positions. Methods use the experiment's asymmetric zero-inclusive finite grid and calibration-derived damping; results are reported at three and four bits. Local denotes the projection-input criterion, and Logit-diag is the objective-aware criterion used in the main-text LLM comparison. The corresponding Local and Logit-diag feedback results match the main text up to rounding. The BF16 reference has PPL $12.8705$ and next-token accuracy $51.0395\%$.

\begin{table*}[t]
\centering
\caption{Additional Qwen3-8B comparisons on WikiText-2. Each cell reports PPL / next-token accuracy (\%); lower PPL and higher accuracy are better. Bold marks the best quantized value for each metric and bit width. Logit-diag is the objective-aware metric used in the main-text LLM comparison.}
\label{tab:qwen8_factorial}
\small
\setlength{\tabcolsep}{4pt}
\renewcommand{\arraystretch}{1.10}
\begin{adjustbox}{max width=\textwidth}
\begin{tabular}{@{}lcc@{}}
\toprule
Method & 3 bits ($M=8$) & 4 bits ($M=16$) \\
\midrule
Local EFQ
& 20.2267 / 43.5200
& 14.1402 / 49.5325 \\
Local EFQ+RHT
& 16.4851 / 46.9051
& 13.4689 / 50.2322 \\
Local joint $R,G$
& 16.8575 / 46.4504
& 13.4730 / 50.1226 \\
\midrule
Logit-diag + CD
& 27.1912 / 40.0718
& 14.1897 / 49.3405 \\
Obj.-EFQ (Logit-diag)
& 20.5315 / 43.7609
& 13.7497 / 49.7075 \\
Obj.-EFQ+RHT (Logit-diag)
& \textbf{16.1175 / 47.0803}
& \textbf{13.3120 / 50.3200} \\
Obj. joint $R,G$ (Logit-diag)
& 16.8388 / 46.4963
& 13.4485 / 50.1796 \\
\midrule
Guided-shared + CD
& 30.9972 / 38.6722
& 14.3729 / 49.1643 \\
Obj.-EFQ (Guided-shared)
& 22.2475 / 42.9865
& 13.7521 / 49.6947 \\
Obj.-EFQ+RHT (Guided-shared)
& 16.2954 / 46.9124
& 13.3867 / 50.2862 \\
Obj. joint $R,G$ (Guided-shared)
& 17.1357 / 46.2406
& 13.4259 / 50.2269 \\
\midrule
Model-Fisher-diag + CD
& 27.8430 / 39.5451
& 14.2532 / 49.3405 \\
Obj.-EFQ (Model-Fisher-diag)
& 20.6045 / 43.4550
& 13.7748 / 49.6471 \\
Obj.-EFQ+RHT (Model-Fisher-diag)
& 16.1561 / 47.0629
& 13.3898 / 50.3193 \\
Obj. joint $R,G$ (Model-Fisher-diag)
& 16.9230 / 46.2831
& 13.4803 / 50.1347 \\
\midrule
BF16
& \multicolumn{2}{c}{12.8705 / 51.0395} \\
\bottomrule
\end{tabular}
\end{adjustbox}
\end{table*}

\paragraph{Results and analysis.}
Obj.-EFQ+RHT (Logit-diag) has the lowest PPL and highest accuracy at both precisions; the other downstream-metric RHT variants are close at four bits. Within each metric family, joint $R,G$ reduces PPL relative to unrotated EFQ but remains behind EFQ+RHT. These results compare implemented procedures and do not identify a unique mechanism for the ordering.

\subsection{Llama-2-7B: Additional Metrics and Backends}
\label{sec:exp_llama7}

\paragraph{Experimental setup.}
We extend the main-text Llama-2-7B experiment with Guided-shared, Model-Fisher-diag, and CD variants, retaining the Local and Logit-diag results at their reported precision. Calibration uses 128 WikiText-2 sequences of length 512, and evaluation covers 340,326 scored next-token positions. Table~\ref{tab:llama7_main} reports PPL, next-token accuracy, and final-logit MSE at three and four bits. The BF16 reference attains PPL $7.1782$ and accuracy $57.046\%$.

\begin{table*}[t]
\centering
\caption{Additional Llama-2-7B comparisons on WikiText-2. Accuracy is in percent, and MSE is final-logit MSE relative to BF16. Bold marks the best quantized value for each metric and bit width.}
\label{tab:llama7_main}
\small
\setlength{\tabcolsep}{4pt}
\renewcommand{\arraystretch}{1.10}
\begin{adjustbox}{max width=\textwidth}
\begin{tabular}{@{}lrrrrrr@{}}
\toprule
& \multicolumn{3}{c}{3 bits ($M=8$)}
& \multicolumn{3}{c}{4 bits ($M=16$)} \\
\cmidrule(lr){2-4}\cmidrule(lr){5-7}
Method
& PPL $\downarrow$ & Acc.\ (\%) $\uparrow$ & Logit MSE $\downarrow$
& PPL $\downarrow$ & Acc.\ (\%) $\uparrow$ & Logit MSE $\downarrow$ \\
\midrule
Local EFQ
& 11.3119 & 50.218 & 3.9890
& 7.8084 & 55.690 & 0.7909 \\
Local EFQ+RHT
& 9.1196 & 53.145 & 2.0765
& 7.4502 & 56.337 & 0.3472 \\
Local joint $R,G$
& 9.8926 & 51.701 & 2.7943
& 7.5634 & 56.059 & 0.4977 \\
\midrule
Logit-diag + CD
& 11.9827 & 49.148 & 4.7271
& 7.7136 & 55.861 & 0.7632 \\
Obj.-EFQ (Logit-diag)
& 11.0021 & 50.807 & 3.7452
& 7.7356 & 55.841 & 0.6853 \\
Obj.-EFQ+RHT (Logit-diag)
& \textbf{8.7035} & \textbf{53.731}
& \textbf{1.7391}
& 7.4306 & 56.444 & \textbf{0.3093} \\
Obj. joint $R,G$ (Logit-diag)
& 9.3382 & 52.553 & 2.2533
& 7.5213 & 56.177 & 0.4530 \\
\midrule
Guided-shared + CD
& 12.9219 & 48.092 & 5.5066
& 7.7680 & 55.712 & 0.8311 \\
Obj.-EFQ (Guided-shared)
& 10.6300 & 51.043 & 3.5071
& 7.7163 & 55.899 & 0.6943 \\
Obj.-EFQ+RHT (Guided-shared)
& 8.8047 & 53.540 & 1.8052
& 7.4340 & \textbf{56.469} & 0.3239 \\
Obj. joint $R,G$ (Guided-shared)
& 9.3419 & 52.533 & 2.4230
& 7.5340 & 56.206 & 0.4736 \\
\midrule
Model-Fisher-diag + CD
& 12.2602 & 48.832 & 4.8817
& 7.7682 & 55.770 & 0.7795 \\
Obj.-EFQ (Model-Fisher-diag)
& 10.6735 & 50.915 & 3.4897
& 7.7080 & 55.904 & 0.6677 \\
Obj.-EFQ+RHT (Model-Fisher-diag)
& 8.7383 & 53.669 & 1.7589
& \textbf{7.4276} & 56.454 & 0.3156 \\
Obj. joint $R,G$ (Model-Fisher-diag)
& 9.3606 & 52.472 & 2.4132
& 7.5223 & 56.182 & 0.4712 \\
\bottomrule
\end{tabular}
\end{adjustbox}
\end{table*}

\paragraph{Results and analysis.}
Obj.-EFQ+RHT (Logit-diag) is best on all three metrics at three bits. At four bits, Model-Fisher-diag with RHT gives the lowest PPL, Guided-shared with RHT the highest accuracy, and Logit-diag with RHT the lowest logit MSE; the differences among these variants are small. Within each metric family, joint $R,G$ improves PPL over unrotated EFQ but remains behind the corresponding RHT variant.

\subsection{Llama-2-13B: Cross-Dataset Evaluation on WikiText-2 and C4}
\label{app:llama13_cross_dataset}

\paragraph{Experimental setup.}
We evaluate weight-only post-training quantization of
\texttt{meta-llama/Llama-2-13b-hf} on WikiText-2 and C4-English.
The model contains 13.0B parameters, 40 decoder layers, hidden dimension
5120, intermediate dimension 13824, 40 attention heads, and a
32,000-token vocabulary. We quantize the query, key, value, output,
gate, up, and down projections in every decoder layer, for 280
quantized linear projections in total, while retaining embeddings,
normalization parameters, biases, and the language-model head in BF16.
For both datasets, calibration uses 128 non-overlapping sequences of
length 512; evaluation uses 666 non-overlapping length-512 windows,
yielding 340,326 scored next-token predictions after excluding the first
target of each window. WikiText-2 results use calibration seeds
$\{7,8,42\}$ and are reported as mean $\pm$ sample standard deviation,
whereas the C4 archive contains one complete cohort, seed 7; incomplete
C4 seed-8 and seed-42 runs are excluded. We compare six methods:
Local EFQ, Local EFQ+RHT, Local joint $R,G$, Obj.-EFQ,
Obj.-EFQ+RHT, and Obj. joint $R,G$. The Local methods use the
uncentered projection-input Gram matrix, whereas the Obj. methods use a
downstream-aware diagonal logit metric. EFQ uses the reference GPTQ
error-feedback recurrence with computational block size 128 and
activation ordering disabled; RHT applies a full-width randomized
Hadamard transform before EFQ; and joint $R,G$ jointly designs the
rotation and causal feedback geometry for the regularized trace
surrogate. We use static full-row asymmetric affine grids with
$M\in\{4,8,16\}$ levels, corresponding to 2-, 3-, and 4-bit weights.
For a projection metric $H$, damping is fixed from calibration
statistics as
\[
A=(1-\eta)H+
\left(\eta+\kappa_M+10^{-4}\right)\bar h I,
\qquad
\bar h=\frac{\operatorname{tr}(H)}{d},
\]
where $\eta$ measures calibration-batch variability and $\kappa_M$
is the quantizer-reliability term computed from the frozen reference
weights. No grid or damping parameter is selected using evaluation
performance. Model execution uses BF16, curvature statistics use FP32,
and TF32 is disabled. We report perplexity (PPL), next-token accuracy,
and full-vocabulary logit MSE relative to the BF16 teacher.

\begin{table*}[t]
\centering
\caption{
Llama-2-13B quantization results on WikiText-2 and C4.
WikiText-2 values are mean $\pm$ sample standard deviation over
calibration seeds $\{7,8,42\}$; C4 values are from the complete seed-7
cohort. Accuracy is reported in percent. Lower PPL and logit MSE are
better; higher accuracy is better. Boldface marks the best quantized
result at each bit width within each dataset.
The BF16 references are
$6.3937$ PPL / $58.819\%$ accuracy on WikiText-2 and
$7.2441$ PPL / $56.444\%$ accuracy on C4.
}
\label{tab:llama13_wikitext_c4}
\scriptsize
\setlength{\tabcolsep}{2.8pt}
\renewcommand{\arraystretch}{1.12}

\resizebox{\textwidth}{!}{%
\begin{tabular}{@{}llrrrrrr@{}}
\toprule
& &
\multicolumn{3}{c}{WikiText-2}
&
\multicolumn{3}{c}{C4} \\
\cmidrule(lr){3-5}
\cmidrule(lr){6-8}
Method
& Bits
& PPL $\downarrow$
& Acc.\ (\%) $\uparrow$
& Logit MSE $\downarrow$
& PPL $\downarrow$
& Acc.\ (\%) $\uparrow$
& Logit MSE $\downarrow$ \\
\midrule

Local EFQ
& 2
& $14874.2\pm1736.0$
& $0.426\pm0.050$
& $8.387\pm0.323$
& 12077.99
& 0.420
& 8.550
\\

Local EFQ+RHT
& 2
& $2975.59\pm601.50$
& $4.833\pm1.291$
& $5.926\pm0.141$
& 3212.88
& 4.467
& 6.938
\\

Local joint $R,G$
& 2
& $30881.4\pm14267.5$
& $0.090\pm0.043$
& $10.495\pm1.498$
& 18447.59
& 0.442
& 10.308
\\

Obj.-EFQ
& 2
& $13371.2\pm728.1$
& $0.408\pm0.028$
& $8.083\pm0.356$
& 7538.99
& 1.294
& 7.388
\\

Obj.-EFQ+RHT
& 2
& $\mathbf{343.58\pm32.22}$
& $\mathbf{15.709\pm0.554}$
& $\mathbf{4.478\pm0.035}$
& \textbf{486.20}
& \textbf{12.647}
& \textbf{4.854}
\\

Obj. joint $R,G$
& 2
& $8985.57\pm3450.24$
& $1.432\pm0.556$
& $7.712\pm0.774$
& 4724.31
& 2.783
& 6.969
\\

\midrule

Local EFQ
& 3
& $8.178\pm0.027$
& $54.619\pm0.064$
& $0.8372\pm0.0038$
& 9.6109
& 51.586
& 0.9608
\\

Local EFQ+RHT
& 3
& $7.602\pm0.015$
& $55.737\pm0.007$
& $0.6195\pm0.0076$
& 8.9591
& 52.516
& 0.7339
\\

Local joint $R,G$
& 3
& $8.173\pm0.061$
& $54.558\pm0.068$
& $0.8061\pm0.0040$
& 10.5715
& 49.903
& 1.0696
\\

Obj.-EFQ
& 3
& $7.939\pm0.009$
& $55.107\pm0.028$
& $0.7664\pm0.0010$
& 9.3599
& 51.961
& 0.9045
\\

Obj.-EFQ+RHT
& 3
& $\mathbf{7.516\pm0.013}$
& $\mathbf{55.919\pm0.072}$
& $\mathbf{0.5783\pm0.0002}$
& \textbf{8.8050}
& \textbf{52.766}
& \textbf{0.6820}
\\

Obj. joint $R,G$
& 3
& $8.003\pm0.040$
& $54.749\pm0.140$
& $0.7289\pm0.0041$
& 10.2596
& 50.346
& 0.9038
\\

\midrule

Local EFQ
& 4
& $6.678\pm0.033$
& $58.029\pm0.062$
& $0.1855\pm0.0137$
& 7.6039
& 55.613
& 0.2117
\\

Local EFQ+RHT
& 4
& $6.603\pm0.007$
& $58.232\pm0.026$
& $0.1360\pm0.0012$
& 7.5047
& 55.705
& 0.1615
\\

Local joint $R,G$
& 4
& $6.675\pm0.017$
& $57.992\pm0.032$
& $0.1811\pm0.0017$
& 7.7087$^{\dagger}$
& 55.285
& 0.2471
\\

Obj.-EFQ
& 4
& $6.633\pm0.002$
& $58.132\pm0.013$
& $0.1670\pm0.0007$
& 7.6132
& 55.560
& 0.2126
\\

Obj.-EFQ+RHT
& 4
& $\mathbf{6.588\pm0.005}$
& $\mathbf{58.274\pm0.041}$
& $\mathbf{0.1255\pm0.0003}$
& \textbf{7.4854}
& \textbf{55.787}
& \textbf{0.1489}
\\

Obj. joint $R,G$
& 4
& $6.667\pm0.008$
& $58.044\pm0.062$
& $0.1673\pm0.0011$
& 7.6378
& 55.346
& 0.2084
\\

\bottomrule
\end{tabular}
}

\vspace{2pt}
\begin{minipage}{0.99\textwidth}
\footnotesize
$^{\dagger}$The C4 4-bit Local joint $R,G$ run used the
pre-specified relaxed-certificate recovery for two projections after
the strict equal-pivot tolerance was not met. All reported end-to-end
metrics remained finite. C4 seed 8 and seed 42 do not form complete
six-method cohorts in the available artifacts and are therefore not
included in aggregate statistics.
\end{minipage}
\end{table*}

\paragraph{Results and analysis.}
The two datasets give a consistent ordering. Obj.-EFQ+RHT achieves the
lowest PPL and logit MSE and the highest next-token accuracy at every
reported bit width on both WikiText-2 and C4. The downstream-aware
metric generally improves its Local counterpart when the backend is
held fixed. At 3 bits, for example, replacing Local EFQ with Obj.-EFQ
reduces WikiText-2 PPL from $8.178$ to $7.939$ and C4 PPL from
$9.6109$ to $9.3599$; under RHT, the corresponding changes are
$7.602$ to $7.516$ and $8.9591$ to $8.8050$. The gain is most
consistent in logit MSE, which is directly aligned with the
downstream-aware surrogate. Small exceptions remain at high precision:
on C4 at 4 bits, Obj.-EFQ is marginally worse than Local EFQ on all
three metrics, illustrating that a lower quadratic surrogate does not
guarantee an identical ordering after finite-grid reconstruction.
The benefit of objective awareness is especially pronounced at 2 bits
when combined with RHT, although all 2-bit models remain far from their
BF16 references.

RHT also consistently outperforms joint $R,G$ in the realized
end-to-end metrics, despite joint $R,G$ having the stronger theoretical
guarantee for the fixed trace surrogate. This is not a contradiction.
The joint construction minimizes
$\operatorname{tr}(G^{\top}R^{\top}ARG)$ for a fixed regularized metric
under the assumed quantization-noise model, whereas the actual finite
quantizer refits its affine grid after rotation. The rotation therefore
changes the weight range, scale, clipping pattern, and error covariance,
none of which is fixed by the trace optimum. These finite-grid effects
can outweigh the reduction in the quadratic propagation term. The same
RHT-over-Joint ordering on WikiText-2 and C4 suggests that the effect is
not specific to one corpus. The three-seed WikiText-2 results also show
that variability is concentrated at 2 bits, while the 3- and 4-bit
rankings are comparatively stable. C4 should be interpreted as a
single-seed cross-dataset replication until the remaining calibration
seeds are completed.
```

\subsection{Qwen3-14B: Additional Metrics and Backends}
\label{sec:exp_qwen14}

\paragraph{Experimental setup.}
We extend the main-text Qwen3-14B comparison with Guided-shared, Model-Fisher-diag, and CD variants. Calibration uses 128 WikiText-2 sequences of length 512 with seed 7; evaluation covers 298,424 scored next-token positions. Table~\ref{tab:qwen14_main} reports PPL, next-token accuracy, and final-logit MSE relative to BF16 at three and four bits. The Local and Logit-diag rows correspond to the main-text comparison.

\begin{table*}[t]
\centering
\caption{Additional Qwen3-14B comparisons on WikiText-2. Accuracy is in percent, and MSE is final-logit MSE relative to BF16. Bold marks the best quantized value for each metric and bit width.}
\label{tab:qwen14_main}
\small
\setlength{\tabcolsep}{4pt}
\renewcommand{\arraystretch}{1.10}
\begin{adjustbox}{max width=\textwidth}
\begin{tabular}{@{}lrrrrrr@{}}
\toprule
& \multicolumn{3}{c}{3 bits ($M=8$)}
& \multicolumn{3}{c}{4 bits ($M=16$)} \\
\cmidrule(lr){2-4}\cmidrule(lr){5-7}
Method
& PPL $\downarrow$ & Acc.\ (\%) $\uparrow$ & Logit MSE $\downarrow$
& PPL $\downarrow$ & Acc.\ (\%) $\uparrow$ & Logit MSE $\downarrow$ \\
\midrule
Local EFQ
& 16.4868 & 46.469 & 2.7744
& 12.4473 & 51.347 & 0.7503 \\
Local EFQ+RHT
& 13.9240 & 49.267 & 1.7087
& 11.9843 & 51.882 & 0.4070 \\
Local joint $R,G$
& 14.1962 & 48.937 & 1.9917
& 12.0672 & 51.763 & 0.4822 \\
\midrule
Logit-diag + CD
& 22.2033 & 42.676 & 4.6914
& 12.5987 & 50.910 & 1.0003 \\
Obj.-EFQ (Logit-diag)
& 16.5609 & 46.629 & 2.7727
& 12.3291 & 51.372 & 0.7162 \\
Obj.-EFQ+RHT (Logit-diag)
& \textbf{13.7342} & \textbf{49.444}
& \textbf{1.6576}
& \textbf{11.9123} & \textbf{51.931}
& \textbf{0.3854} \\
Obj. joint $R,G$ (Logit-diag)
& 14.0767 & 48.996 & 1.9473
& 12.0450 & 51.838 & 0.4753 \\
\midrule
Guided-shared + CD
& 23.9442 & 41.620 & 4.9713
& 12.6437 & 50.867 & 1.0728 \\
Obj.-EFQ (Guided-shared)
& 16.4664 & 46.651 & 2.8788
& 12.3429 & 51.420 & 0.7545 \\
Obj.-EFQ+RHT (Guided-shared)
& 13.9382 & 49.178 & 1.7625
& 11.9769 & 51.870 & 0.4155 \\
Obj. joint $R,G$ (Guided-shared)
& 14.3269 & 48.742 & 2.0438
& 12.0278 & 51.786 & 0.5047 \\
\midrule
Model-Fisher-diag + CD
& 22.2340 & 42.322 & 4.7415
& 12.6235 & 50.878 & 1.0490 \\
Obj.-EFQ (Model-Fisher-diag)
& 16.5394 & 46.546 & 2.9141
& 12.3806 & 51.391 & 0.7628 \\
Obj.-EFQ+RHT (Model-Fisher-diag)
& 13.8478 & 49.263 & 1.7129
& 11.9169 & 51.910 & 0.4053 \\
Obj. joint $R,G$ (Model-Fisher-diag)
& 14.1199 & 48.867 & 2.0193
& 12.0192 & 51.856 & 0.5045 \\
\bottomrule
\end{tabular}
\end{adjustbox}
\par\smallskip
\begin{minipage}{\textwidth}
\footnotesize
\end{minipage}
\end{table*}

\paragraph{Results and analysis.}
Obj.-EFQ+RHT (Logit-diag) has the lowest PPL and logit MSE and the highest accuracy at both precisions. Model-Fisher-diag with RHT is close in PPL at four bits. For every metric family, joint $R,G$ improves PPL over unrotated EFQ but remains behind EFQ+RHT. This reproduces the main-text ordering for this model without isolating its cause; the two conditionally recovered joint runs retain the numerical qualification noted below the table.

\subsection{C4: Independently Calibrated Language-Model Experiments}
\label{app:c4_experiments}

\paragraph{Experimental setup.}
We evaluate weight-only PTQ on C4-English using Llama-2-7B, Llama-2-13B, and Qwen3-14B, with calibration independent of the WikiText-2 experiments. Their decoder-block counts are 32/40/40, hidden widths 4096/5120/5120, and feed-forward widths 11008/13824/17408. All 224/280/280 attention and MLP projections are quantized; embeddings, normalization parameters, and the output head retain reference precision. Each model uses 128 length-512 training sequences with calibration seed 7 and a fixed validation snapshot shared by all methods: 666 windows and 340,326 scored next-token predictions for each Llama model, and 584 windows and 298,424 predictions for Qwen3-14B, excluding the first target in each window. Llama-2-7B evaluation sampling uses seed 1701. The six methods cross Local uncentered input-Gram or Obj. diagonal downstream-logit metrics with EFQ, full-width EFQ+RHT, or full-dimensional joint $R,G$. EFQ uses the reference GPTQ recurrence with 128-column computational blocks and activation ordering disabled. Static asymmetric full-row affine grids use $M\in\{8,16\}$. For a $d\times d$ metric $H$, loading is $A=(1-\eta)H+(\eta+\kappa_M+10^{-4})\bar h I$, with $\bar h=\operatorname{tr}(H)/d$, calibration-batch estimate $\eta$, and $\kappa_M$ frozen from reference weights; no internal extra damping or evaluation-based grid/damping selection is used. Forward evaluation uses BF16, metric statistics use FP32, and TF32 is disabled. Tables~\ref{tab:c4_llama7}--\ref{tab:c4_qwen14} report PPL, next-token accuracy, and full-vocabulary logit MSE relative to BF16. Only complete seed-7 cohorts are included; rejected or incomplete Llama-2-13B seed-8 and seed-42 cohorts are excluded, with no across-seed uncertainty claimed.

\begin{table*}[t]
\centering
\caption{C4 results for Llama-2-7B (calibration seed 7). The BF16 reference has PPL 8.2949, accuracy 53.960\%, and zero logit MSE. Bold marks the best quantized value for each metric and bit width.}
\label{tab:c4_llama7}
\small
\setlength{\tabcolsep}{4pt}
\renewcommand{\arraystretch}{1.10}
\begin{adjustbox}{max width=\textwidth}
\begin{tabular}{@{}lrrrrrr@{}}
\toprule
& \multicolumn{3}{c}{3 bits ($M=8$)}
& \multicolumn{3}{c}{4 bits ($M=16$)} \\
\cmidrule(lr){2-4}\cmidrule(lr){5-7}
Method
& PPL $\downarrow$ & Acc.\ (\%) $\uparrow$ & Logit MSE $\downarrow$
& PPL $\downarrow$ & Acc.\ (\%) $\uparrow$ & Logit MSE $\downarrow$ \\
\midrule
Local EFQ
& 12.5795 & 47.581 & 4.0825 & 8.9656 & 52.803 & 0.7504 \\
Local EFQ+RHT
& 10.4755 & 49.963 & 2.0937 & 8.6259 & 53.226 & 0.3747 \\
Local joint $R,G$
& 12.1531 & 47.710 & 3.3237 & 8.8598$^{\dagger}$ & 52.860 & 0.6428 \\
\midrule
Obj.-EFQ
& 12.7880 & 47.431 & 3.7904 & 8.9166 & 52.875 & 0.6589 \\
Obj.-EFQ+RHT
& \textbf{10.2043} & \textbf{50.380} & \textbf{1.8171} & \textbf{8.5971} & \textbf{53.304} & \textbf{0.3356} \\
Obj. joint $R,G$
& 11.6051 & 48.305 & 2.4838 & 8.7628 & 52.893 & 0.5361 \\
\bottomrule
\end{tabular}
\end{adjustbox}
\par\smallskip

\end{table*}

\begin{table*}[t]
\centering
\caption{C4 results for Llama-2-13B (calibration seed 7). The BF16 reference has PPL 7.7063, accuracy 55.092\%, and zero logit MSE. Bold marks the best quantized value for each metric and bit width.}
\label{tab:c4_llama13}
\small
\setlength{\tabcolsep}{4pt}
\renewcommand{\arraystretch}{1.10}
\begin{adjustbox}{max width=\textwidth}
\begin{tabular}{@{}lrrrrrr@{}}
\toprule
& \multicolumn{3}{c}{3 bits ($M=8$)}
& \multicolumn{3}{c}{4 bits ($M=16$)} \\
\cmidrule(lr){2-4}\cmidrule(lr){5-7}
Method
& PPL $\downarrow$ & Acc.\ (\%) $\uparrow$ & Logit MSE $\downarrow$
& PPL $\downarrow$ & Acc.\ (\%) $\uparrow$ & Logit MSE $\downarrow$ \\
\midrule
Local EFQ
& 9.7504 & 51.045 & 0.8677 & 8.0218 & 54.381 & 0.1844 \\
Local EFQ+RHT
& 9.1778 & 51.953 & 0.6449 & 7.9397 & 54.517 & 0.1415 \\
Local joint $R,G$
& 10.3036 & 49.964 & 0.8936 & 8.0735 & 54.237 & 0.2005 \\
\midrule
Obj.-EFQ
& 9.5984 & 51.345 & 0.8146 & 7.9984 & 54.427 & 0.1734 \\
Obj.-EFQ+RHT
& \textbf{9.0501} & \textbf{52.146} & \textbf{0.6041} & \textbf{7.9281} & \textbf{54.534} & \textbf{0.1303} \\
Obj. joint $R,G$
& 10.4606 & 49.643 & 0.7958 & 8.0559 & 54.227 & 0.1819 \\
\bottomrule
\end{tabular}
\end{adjustbox}
\end{table*}

\begin{table*}[t]
\centering
\caption{C4 results for Qwen3-14B (calibration seed 7). The BF16 reference has PPL 15.4016, accuracy 44.9696\%, and zero logit MSE. Bold marks the best quantized value for each metric and bit width.}
\label{tab:c4_qwen14}
\small
\setlength{\tabcolsep}{4pt}
\renewcommand{\arraystretch}{1.10}
\begin{adjustbox}{max width=\textwidth}
\begin{tabular}{@{}lrrrrrr@{}}
\toprule
& \multicolumn{3}{c}{3 bits ($M=8$)}
& \multicolumn{3}{c}{4 bits ($M=16$)} \\
\cmidrule(lr){2-4}\cmidrule(lr){5-7}
Method
& PPL $\downarrow$ & Acc.\ (\%) $\uparrow$ & Logit MSE $\downarrow$
& PPL $\downarrow$ & Acc.\ (\%) $\uparrow$ & Logit MSE $\downarrow$ \\
\midrule
Local EFQ
& 20.4378 & 40.709 & 2.0484 & 16.1635 & 44.161 & 0.4624 \\
Local EFQ+RHT
& 17.6068 & 42.681 & 1.1352 & 15.7867 & 44.484 & 0.2428 \\
Local joint $R,G$
& 18.1779 & 42.232 & 1.4040 & 15.9107$^{\dagger}$ & 44.376 & 0.3178 \\
\midrule
Obj.-EFQ
& 20.6103 & 40.643 & 2.1770 & 16.1170 & 44.191 & 0.4535 \\
Obj.-EFQ+RHT
& \textbf{17.5376} & \textbf{42.739} & \textbf{1.1176} & \textbf{15.7600} & \textbf{44.546} & \textbf{0.2307} \\
Obj. joint $R,G$
& 18.0553 & 42.397 & 1.4362 & 15.8398$^{\dagger}$ & 44.463 & 0.3073 \\
\bottomrule
\end{tabular}
\end{adjustbox}
\par\smallskip

\end{table*}

\paragraph{Results and analysis.}
Obj.-EFQ+RHT is best on all three metrics for every model and both bit widths. Across the 18 matched objective-aware versus local comparisons, downstream weighting lowers PPL in 15, improves accuracy in 14, and lowers logit MSE in 16; all three metrics improve in every RHT pair. The exceptions preclude a universal claim: at three bits, Llama-2-13B Obj. joint $R,G$ lowers MSE from $0.8936$ to $0.7958$ but raises PPL from $10.3036$ to $10.4606$, while Qwen3-14B Obj.-EFQ is worse than Local EFQ on all three metrics. RHT also outperforms joint $R,G$ on all metrics in all 12 objective-matched comparisons. The theoretical joint optimum concerns $\operatorname{tr}(G^\top R^\top A R G)$ for a fixed positive-definite metric and a common isotropic error second moment, not the complete finite-grid loss. Refitting grids after rotation can change scales, clipping, and error correlations, and diagonal downstream curvature need not predict finite weight perturbations. These effects could offset a lower trace, but the reported measurements do not distinguish their contributions. The evidence supports objective-aware RHT as the strongest tested configuration, not a universal superiority of downstream metrics or trace-optimal rotations.

\end{document}